\documentclass[twocolumn,10pt]{IEEEtran}
\usepackage{amssymb}
\usepackage{theorem,amsbsy,pifont,verbatim,amsfonts,amssymb,color}
\usepackage{xspace,epsfig,graphicx,subfigure,latexsym,fancyhdr,cite}
\usepackage{multirow}
\usepackage{epstopdf}
\usepackage{makecell}
\usepackage{bm}
\usepackage{booktabs}
\usepackage{indentfirst}
\usepackage{amsmath}
\usepackage{diagbox}
\usepackage{enumerate}
\usepackage{float}
\usepackage{url}

\def\bs {\boldsymbol}

\def\0{\boldsymbol 0}
\def\1{\boldsymbol 1}

\def\q{{\boldsymbol q}}

\def\v{{\boldsymbol v}}

\def\w{{\boldsymbol w}}

\def\u{\boldsymbol u}

\newenvironment{proof}{{\it Proof:}}{\hfill $\blacksquare$\par}
\allowdisplaybreaks

\begin{document}

\title{Robust Ellipse Fitting Based on Maximum Correntropy Criterion With Variable Center}

\author{Wei Wang, Gang Wang, \emph{Senior Member, IEEE}, Chenlong Hu, and K. C. Ho, \emph{Fellow, IEEE}
	
\thanks{Wei Wang, Gang Wang, and Chenlong Hu are with the Faculty of Electrical Engineering and Computer Science, Ningbo University, Ningbo 315211, China (e-mail: 2011082136@nbu.edu.cn, wanggang@nbu.edu.cn, huchenlong1223@163.com).}	
\thanks{K. C. Ho is with Electrical Engineering and Computer Science Department, University of Missouri, Columbia, MO 65211, USA (e-mail: hod@missouri.edu).}
\thanks{Corresponding author: Gang Wang.}}

\maketitle
\begin{abstract}
The presence of outliers can significantly degrade the performance of ellipse fitting methods. We develop an ellipse fitting method that is robust to outliers based on the maximum correntropy criterion with variable center (MCC-VC), where a Laplacian kernel is used. For single ellipse fitting, we formulate a non-convex optimization problem to estimate the kernel bandwidth and center and divide it into two subproblems, each estimating one parameter. We design sufficiently accurate convex approximation to each subproblem such that computationally efficient closed-form solutions are obtained. The two subproblems are solved in an alternate manner until convergence is reached. We also investigate coupled ellipses fitting. While there exist multiple ellipses fitting methods that can be used for coupled ellipses fitting, we develop a couple ellipses fitting method by exploiting the special structure. Having unknown association between data points and ellipses, we introduce an association vector for each data point and formulate a non-convex mixed-integer optimization problem to estimate the data associations, which is approximately solved by relaxing it into a second-order cone program. Using the estimated data associations, we extend the proposed method to achieve the final coupled ellipses fitting. The proposed method is shown to have significantly better performance over the existing methods in both simulated data and real images.
\end{abstract}
\section{Introduction}
As a basic function of computer vision, ellipse fitting has been extensively studied. The task of ellipse fitting is to fit a series of data points to an ellipse, which finds wide applications in the fields of aerospace industry, medical imaging, biometrics, pupil recognition, and others {\cite{Rock,glaucoma,gaze,pupil}}. For example, ellipse fitting can detect effectively the rocky and other dangerous areas to ensure a stable and smooth landing of the lunar lander \cite{Rock}. Techniques based on ellipse fitting can accurately detect whether a patient has glaucoma symptoms in medical related fields \cite{glaucoma}. Pupil recognition technology, which has been popular in recent years, also relies on ellipse fitting \cite{gaze}. Moreover, pupil tracking through ellipse fitting has been extensively used in human-computer interactive eye tracker, and has achieved success in consumer electronics industry \cite{pupil}.

In the past several decades, ellipse fitting has attracted a lot of attention and many solutions have been proposed. A traditional method for ellipse fitting is based on the Hough transform (HT) \cite{HT}.
It can achieve high-precision fitting through a voting mechanism in a five dimensional parameter space, but takes a huge computational load.
Owing to this drawback, more computationally efficient methods based on the least squares (LS) approach were proposed \cite{SJ2001,Y2011,directa,Halir1998,directb,directc,directd,ES2006,J2015}. These methods typically work well for clean or simple instances. However, the performance of the LS based methods can degrade a lot in the presence of outliers, which deviate significantly from the underlying ellipse, and appear very often in the extracted data points. As a result, robust ellipse fitting methods that are resilient against the outliers were proposed  \cite{RANSAC,single,hu2021,HGMM}. Early years, some researchers proposed to use a portion of the sample points instead of the whole to resist the interference of outliers in the fitting performance. For instance, Fischler \textit{et al.} proposed a method using  random sample consensus (RANSAC) \cite{RANSAC}, which achieves fitting by estimating a mathematical model from the collection of a random subset of the entire amount of data.

Alternatively, the ellipse fitting problem can be formulated to an optimization problem that can provide robustness.
The papers \cite{single} and \cite{hu2021} solved the ellipse fitting problem by using the maximum correntropy criterion (MCC), where the Gaussian and Laplacian kernels were used, respectively. The MCC method in \cite{single} iteratively solves a number of semidefinite programs (SDPs), and it involves relaxation to the original problem. The relaxation may lead to divergence of the iterations and causing performance loss. By contrast, the MCC method in \cite{hu2021} iteratively solves a set of more computationally efficient second-order cone programs (SOCPs)  \cite{MS1998}, in which no relaxation was introduced. This, in turn, ensures convergence of the iterations. Generally, the MCC ellipse fitting methods have great robustness to the outliers, and thus have good performance in the presence of outliers. 
Recently, Zhao \textit{et al.} \cite{HGMM} proposed the hierarchical Gaussian mixture model (HGMM) for ellipse fitting in noisy, outliers-contained, and occluded scenarios through the Gaussian mixture model (GMM). This method has high robustness against the outliers and noise when the parameters are chosen properly. However, the results can be unsatisfactory when using one particular set of parameters for fitting different ellipses. 

At present, most works in this research address the fitting of single ellipse. However, multiple ellipses with some commonality can be fitted together by exploiting the special structure. An example in practice is the coupled ellipses. Coupled ellipses refer to ellipses that are concentric and have the same rotation angle; their half-short and half-long axes are different but related by a proportional factor. A typical example of coupled ellipses is the inner and outer boundaries of a transmission pipe and the inner and outer edges of an iris image. More applications related to coupled ellipses can be found in \cite{WLS} and the references therein.

Compared with single ellipse fitting, coupled ellipses fitting is more challenging because it has one additional parameter for estimation. To our best knowledge, the available methods for coupled ellipses fitting are very limited in the literature. Ma and Ho \cite{WLS} proposed a weighted least squares (WLS) method for the fitting  problem. The method can reach the Kanatani-Cramer-Rao (KCR) lower bound accuracy without outliers. However, it is sensitive to outliers and the performance degrades significantly when they occur. To improve the robustness against outliers, the work {\cite{hu2021}} applied the MCC based method using the Laplacian kernel and found the solution by alternately solving two subproblems. The results show that it is robust to the outliers even when a large number of them are present.

It has been shown that the MCC is very powerful to solve some signal processing problems, where the data are contaminated by heavy-tailed impulsive noise \cite{noise}. It has also been successfully applied to the ellipse fitting problem \cite{single,hu2021}. However, the existing methods use zero-mean kernel functions, which may not match well with non-zero-mean error distributions, leading to some performance loss. Moreover, the kernel bandwidths in \cite{single} and \cite{hu2021} are computed according to the Silverman's rule \cite{silverman}. The computed kernel bandwidths may not be accurate, which is particularly happening for the case of small number of samples. Moreover, the Silverman's rule is designed for the Gaussian kernel and provides the optimal parameter values  only when the samples are Gaussian distributed \cite{KDE}. As such, Chen \textit{et al.} \cite{B2019} proposed the MCC with variable center (MCC-VC) to improve the performance of MCC, in which both the kernel bandwidth and center can vary to best model the data. MCC-VC is more general than MCC and can be used to handle a broader class of problems.

In this paper, we propose an MCC-VC based ellipse fitting method that is robust against the outliers, in which the Laplacian kernel is used. Similar to \cite{hu2021}, we focus on accurate fitting of the ellipse with the knowledge of the association between the data points and ellipses; hence, strictly speaking, the ``outliers'' here refer to ``pseudo outliers''. In the proposed MCC-VC method, the kernel bandwidth and center are estimated using the available error samples. In the original work proposing MCC-VC \cite{B2019}, the estimation of these two parameters is not well explained. We first develop a new explanation and formulate an optimization problem for the estimation of the kernel bandwidth and center in MCC-VC based on the kernel density estimation (KDE). The proposed optimization problem can be reduced to the same as that in \cite{B2019}. We then propose to solve the optimization problem by dividing it into two subproblems, each estimating one parameter. Rather than using iterative method that requires good initialization, we design sufficiently accurate convex approximation to each subproblem that results in computationally efficient closed-form solution. Specifically, we propose a fourth-order polynomial approximation by applying the Taylor expansion to the objective function when estimating the kernel bandwidth. More importantly, the approximate convex subproblem has a closed-form solution, thereby solving it involves very low computational complexity. The approximation is updated with the kernel bandwidth estimate until convergence. Moreover, for the subproblem of estimating the kernel center, we design a linear programming (LP) problem. The solution of the LP problem turns out to be the median of the available error samples, and thus it is also in closed-form. Using the LP solution as a reference point, we further propose a bisection method for updating the kernel center estimate.

Armed with the estimated kernel bandwidth and center, the estimation of the ellipse parameters reduces to an optimization problem based on the MCC with Laplacian kernel, which can be solved by using the iterative method in \cite{hu2021}. It is worth noting that we formulate a second-order cone (SOC) constraint different from that in \cite{hu2021} to ensure that the fitting curve is elliptical, which enables us to solve the SOCP problem only once in each iteration, instead of twice in the original MCC method \cite{hu2021}. Hence, the new SOC constraint further reduces the computational complexity of solving the MCC problem. The estimated ellipse parameters are used to form new error samples for updating the estimates of the kernel bandwidth and center. The estimation of the kernel bandwidth and center and the estimation of the ellipse parameters are repeated alternatively until convergence.

Since coupled ellipses can be viewed as multiple ellipses with a special structure, the multiple ellipses fitting methods, e.g., those in \cite{HT,RANSAC,HGMM,Lu2020,Prasad2012,TCYBcell} are applicable to coupled ellipses fitting. However, the special structure of coupled ellipses is not exploited in these methods, which may introduce performance loss. Hence, specially developed methods for coupled ellipses fitting are necessary. Existing methods for coupled ellipses fitting \cite{WLS,hu2021} require the prior knowledge of the associations between data points and ellipses, which is not direct or impossible to obtain in practice. To alleviate this issue, we introduce a length 2 association vector, one for each data point composed of 0 and 1, for indicating the association between every data point to either ellipse, and formulate a mixed integer optimization problem to jointly estimate the ellipse parameters and the association vector. It is commonly known that the mixed integer problem is very difficult to solve. To make the problem tractable, we relax the association vector to a probability vector, yielding a convex SOCP. An estimate of the association vector can be deduced from the SOCP solution. Using the estimated association vectors, the proposed MCC-VC fitting method is extended to coupled ellipses fitting. It is worth noting that the incorrectly associated data points are treated as outliers during the fitting, implying the fitting method needs to be robust to the outliers.
The contributions are summarized as follows, including:
\begin{itemize}	
	\item[1)] We propose a robust ellipse fitting method based on MCC-VC, in which the kernel bandwidth and center are estimated by a well explained optimization problem. Moreover, this problem is efficiently solved by accurate convex approximations.
	\item[2)] We formulate a new SOC constraint to guarantee the fitted curve is an ellipse, which greatly improves the computational efficiency of the proposed method. 
	\item[3)] We propose a data association method for the coupled ellipses fitting problem when the associations between data points and ellipses are not known. With the estimated data associations, we extend the proposed MCC-VC fitting method to achieve the final coupled ellipses fitting.
\end{itemize}
  
The organization of the paper is as follows. Section \ref{s2} gives the measurement models for the single ellipse, and coupled ellipses with unknown data association. In Section \ref{s3}, we present the optimization problems for single ellipse fitting based on MCC-VC, and derive the iterative method to solve the optimization problems. Section \ref{s4} develops the data association method and presents the MCC-VC method for coupled ellipses fitting. Section \ref{s5} demonstrates the performance of the proposed fitting method by several experiments using both simulated data and real images. Section \ref{s6} concludes the paper.
We shall use the following notations throughout the paper. Vectors and matrices are represented by boldface lowercase and boldface uppercase letters, respectively. $(\ast)^o$ is the true value of $(\ast)$. $\mathbb{E}[\ast]$ is the mathematical expectation of vector $(\ast)$.  $(\ast)_{(k)}$ is the $k$th element in vector $(\ast)$. $|\ast|$ and $\|\ast\|$ are the $\ell_1$-norm and $\ell_2$-norm, respectively. Other mathematical symbols are defined when they first appear.

\section{System Models}\label{s2}

\subsection{Single Ellipse}

In the 2-D Euclidean space, an ellipse can be uniquely determined by five parameters: the center $(g,h)$, half-long axis $a$, half-short axis $b$ and counter-clockwise rotation angle $\theta$. The ellipse constructed by these five parameters is \cite{hu2021}
\begin{small}\begin{align}\label{e1}
	&\frac{[(x^o-g)\cos\theta+(y^o-h)\sin\theta]^2}{a^2}+\nonumber\\
	&\frac{[-(x^o-g)\sin\theta+(y^o-h)\cos\theta]^2}{b^2}=1,
\end{align}\end{small}
where $(x^o,y^o)$ denotes a regular point on an ellipse.

The available data points, possibly having outliers, can be modeled by
\begin{small}
\begin{align}\label{e2a}
	&{x}_i={x}_i^o+v_i^x+n_i^x, \; {y}_i={y}_i^o+v_i^y+n_i^y,
\end{align}	
\end{small}	
where $n_i^x$ and $n_i^y$ are the measurement noise, and $v_i^x$ and $v_i^y$ are zero if the pair is a normal noise only contaminated point and have larger values if it is an outlier.	

Substituting (\ref{e2a}) into (\ref{e1}) and after simple manipulations, we can rewrite (\ref{e1}) as the following implicit second-order polynomial equation \cite{directa}, \cite{single,hu2021}:

\begin{small}
\begin{align}\label{e3}
	Ax_i^2+Bx_iy_i+Cy_i^2+Dx_i+Ey_i+F={ \delta_i},\; i=1,\ldots,N,
\end{align}	
\end{small}	
where $\delta_i$ is the measurement error induced by noise and outliers. For the sake of simplicity, we introduce the following vectors:
\begin{small}
\begin{align}\label{e4}
	&\v=[A,B,C,D,E,F]^T, \;\q=[g,h,a,b,\theta]^T,\nonumber\\
	&\u_i=[x_i^2,x_iy_i,y_i^2,x_i,y_i,1]^T.
\end{align}	
\end{small}	
The model (\ref{e3}) can be expressed in the vector form 
\begin{small}
\begin{align}\label{e4a}
	\v^T\u_i={ \delta_i},\; i=1,\ldots,N.
\end{align}	
\end{small}	
To guarantee the model (\ref{e3}) represents an ellipse but not a hyperbola, the condition $B^2-4AC<0$, i.e., $v_{(2)}^2-4v_{(1)}v_{(3)}<0$, must be satisfied. The objective is to solve the ellipse fitting problem to obtain an optimal estimate of $\v$ with the condition $v_{(2)}^2-4v_{(1)}v_{(3)}<0$ being satisfied. From the estimate of $\v$, we can obtain the estimate of the parameters of the ellipse $\q$ by the common conversion formulas \cite{hu2021}.

\subsection{Coupled Ellipses}

The equations for the outer and inner coupled ellipses can be expressed as \cite{WLS}
\begin{small}
\begin{align}\label{e8} 
	&\frac{[(x^o-g)\cos\theta+(y^o-h)\sin\theta]^2}{a_j^2}+\nonumber\\
	&\frac{[-(x^o-g)\sin\theta+(y^o-h)\cos\theta]^2}{b_j^2}=1,
\end{align}	
\end{small}	
where $a_j$ and $b_j$ for $j=1,2$ are the half-long and half-short axes, respectively. Their lengths satisfy $a_2={\mu}a_1$ and $b_2={\mu}b_1$ with $\mu\in(0,1)$ being the proportional factor.

To seek the relationship between the parameters of inner ellipse and those of the outer, we first multiply an arbitrary positive factor $\beta$ to both sides of (\ref{e8}):
\begin{small}
\begin{align}\label{e9}
	&\beta\bigg(\frac{[(x^o-g)\cos\theta+(y^o-h)\sin\theta]^2}{a_j^2}+\nonumber\\
	&\frac{[-(x^o-g)\sin\theta+(y^o-h)\cos\theta]^2}{b_j^2}\bigg)=\beta,\;j=1,2.
\end{align}	
\end{small}	

Assume that we have collected $N$ noisy data points $(x_i,y_i)$, $i=1,2,\cdots,N$, where the associations between the data points and the ellipses are not known. After substituting $(x_i,y_i)$ to both equations of the outer and inner ellipses in  (\ref{e9}), the model equations become \cite{hu2021}
\begin{small}
\begin{subequations}\label{e12}
	\begin{align}
		&A{x_i}^2+Bx_iy_i+C{y_i}^2+Dx_i+Ey_i+F=\delta_{i,1},\label{e12a}\\
		&A{x_i}^2+Bx_iy_i+C{y_i}^2+Dx_i+Ey_i+F+\eta=\delta_{i,2},\label{e12b}
	\end{align}
\end{subequations}	
\end{small}	
where $\eta=\beta(1-\mu^2)$ and $\delta_{i,j}$ for $j=1,2$ are the equation errors of the $i$th data point $(x_i,y_i)$ corresponding to the outer and inner ellipses. Typically, $|\delta_{i,1}|<|\delta_{i,2}|$ if the $i$th point belongs to the outer ellipse, and $|\delta_{i,1}|>|\delta_{i,2}|$ otherwise.

The equations in (\ref{e12a}) and (\ref{e12b}) can be further represented by the following vector form:
\begin{small}
\begin{align}\label{e39}
	[\v^T\u_i\;\;\v^T\u_i]^T+\bm{\tau}&=\bm{\delta}_i,\; i=1,\ldots,N,
\end{align}	
\end{small}	
where
\begin{small}
\begin{align}\label{e40}
	&\v=[A,B,C,D,E,F]^T, \; \bm{\tau}=[0\;\;\eta]^T\nonumber\\
	&\u_i=[x_i^2,x_iy_i,y_i^2,x_i,y_i,1]^T,	{\bm{\delta}}_i=[\delta_{i,1}\;\;\delta_{i,2}]^T. 
\end{align}	
\end{small}	
To explicitly express the model equation for coupled ellipses fitting that includes data association, we introduce an association vector $\bm{\phi}_i$ to represent the association between the ellipse and the $i$th noisy data point. It will take the value of $\bm{\phi}_i=[1,0]^T$ if the $i$th data point is associated with the outer ellipse and $\bm{\phi}_i=[0,1]^T$ if it is associated with the inner ellipse. $\bm{\phi}_i$ are not known and to be estimated. The model for coupled ellipses fitting after data association is obtained by multiplying the association vector $\bm{\phi}_i$ to both sides of (\ref{e39}):
\begin{small}
\begin{align}\label{e41}
	\bm{\phi}^T_i{\bm{\delta}}_i=\v^T\u_i+\bm{\phi}^T_i\bm{\tau},\; i=1,\ldots,N,
\end{align}	
\end{small}	
where the left term is the model error of the $i$th data point.

\section{Single Ellipse Fitting Based on MCC-VC}\label{s3}
In this section, we propose a robust formulation for single ellipse fitting by  MCC-VC, to provide better robustness against the outliers. Moreover, we propose a more efficient procedure than the one in \cite{hu2021}, to guarantee the condition for forming an ellipse, i.e., $v_{(2)}^2-4v_{(1)}v_{(3)}<0$, is fulfilled.

\subsection{MCC-VC With Laplacian Kernel}

%
MCC can be utilized for robust estimation of unknowns. According to \cite{hu2021}, the unknown vector $\bs \alpha$, which is related to the error vector $\delta$, can be estimated using the MCC by 
\begin{small}
\begin{align}\label{e19}
	\max_{\bs \alpha} \; \mathbb{E}[\kappa_{\sigma}(\delta(\bs \alpha))],
\end{align}	
\end{small}	
where $\kappa_{\sigma}(\bullet)$ is the kernel function and $\sigma$ denotes the kernel bandwidth.
The value of the kernel bandwidth $\sigma$ is usually determined according to the Silverman's rule \cite{silverman} when the kernel is Gaussian, which may not be accurate especially when the number of samples used for evaluating the expectation is small or the kernel function is not Gaussian. For this reason, we propose an MCC with adaptive kernel bandwidth $\sigma$, which involves KDE.


In statistics, KDE can be considered as using a non-parametric approach to estimate the probability density function of a random variable. Given a finite number of data samples, KDE is a fundamental data smoothing problem. Specifically, let $\{\delta_i\}_{i=1}^N$ be the independent and identically distributed samples drawn from some univariate distribution of an unknown density $p$. The kernel density estimator $\hat{p}_{\sigma}(\delta)$ for any given point $\delta$ can be expressed as
\begin{small}
\begin{align}\label{e23ab}
	\hat{p}_{\sigma}(\delta)=\frac{1}{N}\sum_{i=1}^N\kappa_{\sigma}(\delta-\delta_i).
\end{align}	
\end{small}	
\;\; In KDE, the kernel bandwidth estimation is key to the performance.  The commonly used criterion for selecting $\sigma$ is to minimize the mean integrated squared error (MISE) \cite{MISE}:
\begin{align}\label{e23aa}
	{\rm MISE}(\sigma)=\mathbb{E}[\int(\hat{p}_{\sigma}(\delta)-p(\delta))^{2}d\delta],
\end{align}
where $p(\delta)$ is the probability density function (PDF) of the true error distribution. However, minimizing the MISE is generally not feasible since the true PDF of the error is not available. To make the problem tractable, we make approximations in the following. 

Since we only have one set of samples, we first remove the expectation operation, i.e., we minimize the integrated squared error (ISE) instead of the MISE. It is not difficult to observe from (\ref{e23ab}) and (\ref{e23aa}) that minimizing ISE is still an intractable problem owing the square of the summation. Thus, we further approximate $\hat{p}_{\sigma}(\delta)$ by using one sample, i.e., $\label{e23ac}
	\hat{p}_{\sigma}(\delta)\approx \kappa_{\sigma}(\delta-c)$,
where $c$ is the an unknown representative sample. $c$ is also known as the center of the kernel function \cite{B2019}.

Finally, we seek optimal $c$ and $\sigma$ by minimizing the ISE. The term $\int (p(\delta))^2d\delta$ is independent of $c$ and $\sigma$, we have
\begin{small}
\begin{align}\label{e23ad}
	\begin{aligned}
		\left(c^{*},\sigma^{*} \right)&=\arg \min_{c,\sigma }  \int\left[\kappa_{\sigma}(\delta-c)-p(\delta)\right]^{2} d \delta \\
		\quad&=\arg \min_{c,\sigma }\left\{\int\left[\kappa_{\sigma}(\delta-c)\right]^{2} d \delta-2\mathbb{E}\left[\kappa_{\sigma}(\delta-c)\right]\right\}. \\
	\end{aligned}
\end{align}	
\end{small}	
\;\; The mathematical expectation in problem (\ref{e23ad}) is approximated by sample averaging over the $N$ samples $\{\delta_i\}_{i=1}^N$.
In this paper, the kernel function is Laplacian for its robustness to outliers. Substituting the Laplacian kernel function $\kappa_{\sigma}({\delta}-c)=\frac{1}{2\sigma}e^{-\frac{\vert{{\delta}-c}\vert}{\sigma}}$ into problem (\ref{e23ad}) gives 
\begin{small}
\begin{align}\label{e23ae}
	\begin{aligned}
		&\left( c^{*},\sigma^{*}\right)=\arg \min_{c,\sigma } \left\{\frac{1}{4 \sigma}-\frac{1}{N\sigma}\sum_{i=1}^N{e^{-\frac{|\delta_i-c|}{\sigma}}}\right\}.
	\end{aligned}
\end{align}	
\end{small}	

It is worth noting that problem (\ref{e23ae}) is obtained differently from \cite{B2019} although the final form is exactly the same as that in \cite{B2019} when the Laplacian kernel function is used. 

With the estimated $c^{*}$ and $\sigma^{*}$ by problem (\ref{e23ae}), MCC follows to estimate the unknown parameters in the model by solving the problem 
\begin{small}
\begin{align}\label{e21}
	&\arg\min_{\bs \alpha}\;-\frac{1}{N}\sum_{i=1}^N\kappa_{\sigma^{*}}( \delta_{i}(\bs \alpha)-c^{*})\nonumber\\
	=&\arg \min_{\bs \alpha}\; -\frac{1}{\sigma^{*}}\sum_{i=1}^Ne^{-\frac{\vert{\delta_{i}(\bs \alpha)-c^{*}}\vert}{\sigma^{*}}}.
\end{align}	
\end{small}	 
where $2N$ is discarded as a constant irrelevant to the optimization variable. To summarize, problems (\ref{e23ae}) and (\ref{e21}) are solved in the MCC-VC method.


\subsection{Single Ellipse Fitting Based on MCC-VC}

In the ellipse fitting problem, $\v$ is the unknown vector to be estimated. We shall deduce an optimization problem to estimate $\v$ based on MCC-VC. 


It is seen from (\ref{e4a}) that $\delta_i$ is related to the unknown vector $\v$. Thus, we express $\delta_i$ as  $\delta_i(\v)=\v^{T}\u_i$ for $i=1,\ldots, N$. Using an estimate of $\v$, denoted by $\hat{\v}$, we can construct the samples $\delta_i(\hat{\v})$. Problem (\ref{e23ae}) becomes
%
\begin{small}
\begin{align}\label{e23ag}
	\begin{aligned}
		&\left(\hat{c},\hat{\sigma}\right)=\arg \min_{ c,\sigma} \left\{\frac{1}{4 \sigma}-\frac{1}{N\sigma}\sum_{i=1}^N{e^{-\frac{|{\hat{\v}}^T{\u_i-c}|}{\sigma}}}\right\}.
	\end{aligned}
\end{align}	
\end{small}	
With the estimated $c$ and $\sigma$, denoted by $\hat{c}$ and $\hat{\sigma}$, the ellipse parameter vector can be estimated by solving
\begin{small}
\begin{align}\label{e23}
	\hat{\v}=\arg \min_{\v}& \; -\frac{1}{\hat{\sigma}}\sum_{i=1}^N{e^{-\frac{\vert{{\v}^T{\u_i-\hat{c}}}\vert}{\hat{\sigma}}}}\nonumber\\
	{\rm s.t.}& \;\;v_{(2)}^2-4v_{(1)}v_{(3)}<0,
\end{align}\end{small}	
where the condition $v_{(2)}^2-4v_{(1)}v_{(3)}<0$ is included as a constraint to ensure the solution is an ellipse. 

In the following, we develop specific methods to solve problems (\ref{e23ag}) and (\ref{e23}) iteratively for estimating the unknown parameters $c$, $\sigma$, and $\v$ in an alternate manner.

\subsubsection{Estimation of the Kernel Bandwidth and Center}\label{B-1}

Joint estimation of the kernel bandwidth and center in problem (\ref{e23ag}) may lead to local convergence owing the non-convex nature of the problem. To avoid the local convergence issue, we propose to estimate the two parameters by dividing problem (\ref{e23ag}) into two subproblems, with one parameter estimated by one subproblem. The subproblems are not solved using the routine gradient based methods, such as the gradient decent method and Newton's method. Instead, we design a sufficiently accurate convex problem for each subproblem, and the solution of the convex problem is used as a starting point to obtain the optimal solution of the corresponding subproblem by local search. Since the designed convex problems are sufficiently accurate to provide good starting points, we expect to obtain the optimal solutions of the original non-convex subproblems, although global convergence is not guaranteed.  


\textit{i) Estimation of the kernel bandwidth}

We first present a procedure to estimate the kernel bandwidth $\sigma$ when fixing $\v$ and $c$ to their estimates $\hat  \v$ and $\hat  c$ from the previous iteration. For notational simplicity, let $\hat  \delta_i= \hat  \v^T \u_i$ for $i=1,\ldots, N$. Using $\hat \delta_i$ and $\hat c$, we can estimate $\sigma$ by solving the following problem
\begin{small}
\begin{align}\label{e23a}
	\min_{\sigma}\; &\left\{\frac{1}{4\sigma}-\frac{1}{N\sigma}\sum_{i=1}^N{e^{-\frac{|\hat  \delta_i-\hat  c|}{\sigma}}}\right\}.
\end{align}	
\end{small}	

By letting $r=\frac{1}{\sigma}$, problem (\ref{e23a}) can be written into an optimization problem with $r$ as variable:
\begin{small}
\begin{align}\label{e23b}
	\min_{r}\; \left\{h(r):=\frac{r}{4}-\frac{r}{N}\sum^{N}_{i=1}e^{{-|{\hat{\delta_{i}}-\hat{c}}|}r}\right\}.
\end{align}	
\end{small}	
\;\; Obviously, once the optimal solution of problem (\ref{e23b}) is obtained, the optimal solution of problem (\ref{e23a}) is also readily available.

By computing the second-order derivative of $h(r)$ \cite{convex}, it can be proven that problem (\ref{e23b}) is non-convex, which may lead to possible local convergence if an iterative algorithm is used to solve it. We shall approximate (\ref{e23b}) as a sufficiently accurate convex problem through truncating the Taylor expansion. For brevity, we denote $\hat{a}_{i}=|\hat{\delta_{i}}-\hat{c}|$ for $i=1,\ldots, N$. By introducing a known positive constant $r_0$, Appendix \ref{app2} shows that $h(r)$ can be approximated by the following fourth order polynomial function, i.e.,
\begin{small}
\begin{align}\label{e30b}
	h(r)&\approx f(r)=\frac{b_4}{6}(r-r_0)^4-\frac{b_3}{2}(r-r_0)^3+b_2(r-r_0)^2\nonumber\\
	&+(b_2r_0-b_1+\frac{1}{4})(r-r_0)+(\frac{1}{4}r_0-b_1r_0),
\end{align}	
\end{small}	
where $b_1$, $b_2$, $b_3$, and $b_4$ are defined in Appendix \ref{app2}. The approximation is typically accurate around $r_0$.


It is straightforward to see that the optimal solution of problem (\ref{e23a}) can be obtained by sequentially minimizing $f(r)$. In particular, starting from $r_0=0$, we can minimize $f(r)$ to obtain an updated $r_0$, then we minimize $f(r)$ again using the updated $r_0$ and repeat the process until convergence. 


A close observation to the problem of minimizing $f(r)$ reveals that it has a closed-form solution. According to the Karush-Kuhn-Tucker (KKT) condition, the solution of minimizing $f(r)$ can be obtained by solving the univariate cubic equation $f'(r)=0$, whose real root has a closed-form expression. Appendix \ref{app3} shows the existence, uniqueness, and the expression of the closed-form solution. Note that the closed-form solution implies a lower computational complexity than the gradient based methods to solve problem (\ref{e23a}). Finally, the optimal estimation of $\sigma$ can be obtained by taking the reciprocal of $r$.


\textit{ii) Estimation of the kernel center}

We will present another estimation procedure to estimate the kernel center $c$ when fixing bandwidth $\sigma$ to its estimates $\hat \sigma$ from the previous iteration. Problem (\ref{e23ae}) becomes
\begin{small}
\begin{align}\label{e33a}
	\min_{c}\;-\sum_{i=1}^Ne^{-\frac{\vert{{\hat \delta_{i}}-c}\vert}{\hat \sigma}},
\end{align}	
\end{small}	 
which can be used to obtain the kernel center $c$. 

However, problem (\ref{e33a}) is non-convex owing to the non-convex objective function. To simplify the problem, we shall approximate the objective function of problem (\ref{e33a}) by a convex function. Specifically, we keep only the first-order Taylor expansion of $e^{-\frac{\vert{{\hat \delta_{i}}-c}\vert}{\hat \sigma}}$, i.e., $e^{-\frac{\vert{{\hat \delta_{i}}-c}\vert}{\hat \sigma}}\approx 1-\frac{|{{\hat \delta_{i}}-c}|}{\hat \sigma}.$

We can approximate problem (\ref{e33a}) as an LP problem
\begin{small}
\begin{align}\label{e33}
	\min_{c}\;-\sum_{i=1}^N\left({1-\frac{|{{\hat \delta_{i}}-c}|}{\hat \sigma}}\right)=\min_c \;\sum_{i=1}^N|\hat \delta_{i}-c|.
\end{align}	
\end{small}	
\;\; In fact, problem (\ref{e33}) is the maximum likelihood estimation of the center of the Laplacian distribution from the samples that follow Laplacian distribution \cite{median}, and its solution is the median of the samples. Hence, the solution would be a reasonably approximation even when the samples are not Laplacian distributed, although it is not optimal to the original problem (\ref{e33a}). On the other hand, the optimal solution of problem (\ref{e33a}) can be obtained through a simple procedure using the LP solution as the starting point. Noting that the objective of problem (\ref{e33a}) is non-differentiable, the gradient based method is not applicable. We propose a simple bisection method to solve problem (\ref{e33a}). To this end, we first determine an appropriate interval of $c$ for the bisection method, containing the LP solution and the sample around the LP solution having smallest objective value, and then we perform the bisection method. 
Suppose that the LP solution is accurate enough, the optimal solution of problem (\ref{e33a}) can be obtained.

\subsubsection{Estimation of the Ellipse Parameters}\label{B-3}
When keeping $c$ and $\sigma$ to the values $\hat {c}$ and $ \hat  \sigma$, the MCC-VC problem reduces to the following MCC problem:
\begin{small}
\begin{align}\label{e34}
	&\min_{\v}\;-\sum_{i=1}^N{e^{-\frac{\vert{{\v}^T{\u_i-\hat c}}\vert}{\hat  \sigma}}}, \;{\rm s.t.} \;v_{(2)}^2-4v_{(1)}v_{(3)}<0,
\end{align}	\end{small}	
where the constant scaling parameter ${1}/{\hat \sigma}$ has been dropped. The previous work \cite{hu2021} presented an method in solving problem (\ref{e34}). Specifically, the solution to (\ref{e34}) is obtained by solving the following two subproblems iteratively: 

\textit{Subproblem 1:} Assuming that an estimate of the weight vector $\w$ is available, denoted by $\bar{\w}$, the first subproblem is:
\begin{small} 
	\begin{subequations}\label{e35}
	\begin{align}
		\min_{\v,\zeta_i}\;&\sum_{i=1}^N{(-\bar{w}_i{\zeta_i})} \nonumber\\
		{\rm s.t.}\;&\vert{{\v}^T{\u_i}-\hat  c}\vert\leq \zeta_i,\; i=1,\ldots,N,\label{e35-1}\\
		&v_{(2)}^2-4v_{(1)}v_{(3)}<0.\label{e35-2}
	\end{align}
\end{subequations}	
\end{small}	
\;\; \textit{Subproblem 2:} If an estimate of $\v$, denoted by $\bar{\v}$, is available, we can obtain the optimal estimate of the weight vector ${\w}=[{w}_1,\ldots,{w}_N]^T$ by the property of convex conjugate functions: ${w}_i=-e^{-\frac{\vert{\bar{\v}^T{\u_i-\hat  c}}\vert}{\hat  \sigma}}$.

The solution method typically takes the ``$<$'' as ``$\leq$'' when solving Subproblem 1. A procedure was proposed to guarantee the sign ``$<$'' in \cite{hu2021}, where an SOCP problem was solved twice in one iteration. Different from \cite{hu2021}, we here propose a more efficient procedure for  the condition $v_{(2)}^2-4v_{(1)}v_{(3)}<0$ to be fulfilled. Specifically, we replace the constraint in (\ref{e35-2}) by $ v_{(2)}^2+\varepsilon^2 \leq 4v_{(1)}v_{(3)}
$, where $\varepsilon$ is an arbitrary constant. 

It is straightforward to write the constraint  (\ref{e35-2}) as the following second-order cone constraint:
\begin{small}
\begin{align}\label{e37}
	\left\|\left[
	v_{(2)},\varepsilon,v_{(1)}-v_{(3)}
	\right]^{T}\right\|\leq v_{(1)}+v_{(3)}.
\end{align}	
\end{small}	
Replacing the constraint in (\ref{e35-2}) by that in (\ref{e37}), Subproblem 1 becomes the following SOCP:
\begin{small}
\begin{align}\label{e38}
	\min_{\v,\zeta_i}\;\;&\sum_{i=1}^N{(-\bar{w}_i{\zeta_i})}\nonumber\\
	{\rm s.t.}\;\;&\vert{{\v}^T{\u_i}-\hat  c}\vert\leq\zeta_i,\; i=1,\ldots,N,\;\;\mbox{(\ref{e37})}.
\end{align}	
\end{small}	
\;\; Different from \cite{hu2021}, in which an SOCP problem were solved twice in one iteration, the procedure proposed in this work needs the SOCP problem (\ref{e38}) solved only once. 


After obtaining the estimate of $\v$ by solving problem (\ref{e34}), the kernel center and bandwidth estimates can be updated using the procedure in Section \ref{B-1}. The process repeats until the stopping criterion is reached. From the estimate of $\v$, one can recover the estimate of $\q$ according to the relationship between $\v$ and $\q$ \cite{hu2021}.

The entire MCC-VC method for single ellipse fitting is given in Algorithm 1. It is worth noting that the MCC fitting method in \cite{hu2021} cannot fit the ellipse correctly in some cases owing to the predefined kernel center and bandwidth. Therefore, a clustering technique is used in \cite{hu2021} to detection the failed fitting. Once the failed fitting is detected, the iterations will be restarted by setting a different initialization, which obviously will increase the computational overload. By contrast, the proposed MCC-VC method has very rare failed fittings, and the clustering is generally not needed. Hence, the proposed method is easier to use. 


\begin{small}
\begin{table}[htb]
	\centering
	\begin{minipage}{.5\textwidth}
		\begin{tabular}{l}\toprule
			
			\textbf{Algorithm 1:} The MCC-VC Method for Single Ellipse Fitting\\\midrule
			\textbf{Input}:\\
			\qquad $c^0=0$: initial kernel center; \;$\bar{\w}^0=-1/N$: initial weights;\\
			\qquad\{$\u_i$\}: collected data points; \;$L$: maximum number of iterations;\\
			\qquad$\varepsilon$: constant to guarantee $v_{(2)}^2-4v_{(1)}v_{(3)}<0$;\\
			\textbf{Steps}:\\
			\qquad 0: Solve problem (\ref{e38}) to obtain an initial estimate $\hat{\v}^0$. \\
			\qquad Let $\hat{\delta}_i^{0}=(\hat{\v}^{0})^T\u_i$ and solve problem (\ref{e23a}) to obtain an initial\\ 
			\qquad estimate of $\hat \sigma^0$, where $\hat\delta_{i}=\hat{\delta}_i^0$ and $\hat c=c^0$; \\ 
			\quad \textbf{for}\;$\ell=1:L$\\
			\qquad 1: Solve problem (\ref{e34}) to obtain $\hat{\v}^{\ell}$ and compute $\hat \delta_{i}^{\ell}=(\hat{\v}^{\ell})^T\u_i$;  \\
			\qquad 2: \textit{(i)} Solve problem (\ref{e33a}) to obtain the estimate of $c$,  $\hat c^{\ell}$; \\
			\qquad\quad \textit{(ii)}  Solve problem (\ref{e23a}) to obtain the estimate of $\sigma$,  $\hat \sigma^{\ell}$; \\
			\qquad 3: If $||g(\hat{\v}^{\ell},\hat c^{\ell},\hat \sigma^{\ell})-g(\hat{\v}^{\ell-1},\hat c^{\ell-1},\hat \sigma^{\ell-1})||<10^{-5}$, where \\ \qquad  $g(\v,c,\sigma)\stackrel{\vartriangle}{=}-\frac{1}{\sigma}\sum_{i=1}^N{e^{-\frac{\vert{{\v}^T{\u_i-c}}\vert}{\sigma}}}$ 
			or $\ell=L$, break.\\
			\quad \textbf{end\;for} \\
			\qquad 4: Obtain the the vector $\hat{\q}=[\hat{g},\hat{h},\hat{a},\hat{b},\hat{\theta}]^T$ using $\hat{\v}^{\ell}$, \\
			\qquad according to the relationship between $\v$ and $\q$.\\
			\textbf{Output}: The ellipse parameters $[\hat{g},\hat{h},\hat{a},\hat{b},\hat{\theta}]^T$.\\
			\bottomrule
		\end{tabular} 
	\end{minipage}
\end{table}	
\end{small}

\section{Coupled Ellipses Fitting With Unknown Data Association}\label{s4}

In this section, we investigate the coupled ellipses fitting problem, where the associations between the ellipses and the observed noisy data points are not available, i.e., we do not know which data point belongs to which ellipse. To overcome this difficulty, we propose a practical approach consisting of two steps: Step 1: Associations of the data points to the ellipses; Step 2: Fitting of the associated data points to the coupled ellipses. To be more specific, the task of Step 1 is to estimate the association vectors  $\bm{\phi}_i$ for $i=1,\ldots,N$, and Step 2 extends the proposed MCC-VC method to coupled ellipses fitting by using the estimated association vectors in Step 1. In Step 2,  the incorrectly associated data points in Step 1 are treated as outliers.

\subsection{Data Association}
According to (\ref{e41}) and imposing the condition $v_{(2)}^{2}+\varepsilon^2-4v_{(1)}v_{(3)}<0$ to ensure the result is an ellipse, we formulate the following optimization problem to estimate the association vectors, where the coupled ellipses parameters are estimated in conjunction as well:
\begin{small}
\begin{align}\label{e42}
	\min_{\{\bm{\phi}_i\in\prod^{2\times1}\},\;\v,\bm{\tau}}\;&\sum_{i=1}^N{\vert\v^T{\u_i}+\bm{\phi}^T_i\bm{\tau}\vert}\nonumber\\
	{\rm s.t.}\quad\;& \tau_{(1)} = 0,\;\mbox{(\ref{e37})},
\end{align}	
\end{small}
where $\prod^{2\times1}$ is the set of all possible $2\times1$ association vectors.

Problem (\ref{e42}) is very difficult to solve owing to integer variables and the inner product of the unknown vector $\bm{\phi}_i$ and $\bm{\tau}$. It should be emphasized that the main purpose of problem (\ref{e42}) is to estimate the association vectors but not for coupled ellipses fitting, because the final coupled ellipses will be obtained in the second step. Keeping this in mind, we may still be able to correctly estimate the association vectors by making some approximations, even though the ellipse parameters may not be accurate owing to these approximations. The idea is to make the fitted inner and outer ellipses be located between the true inner and outer ellipses, which will generate correct associations for most noisy data points. To this end, we set $\eta$ that is defined below (\ref{e12}) to unity such that $\bm{\tau}$ shown in (\ref{e40})  is a constant vector. By doing so, the inner product of two unknown vectors becomes the product with only one unknown vector $\bm{\phi}_i$ and problem  (\ref{e42}) is partially simplified. 

The value of $\varepsilon$ in (\ref{e42}) cannot be arbitrarily chosen any more since $\eta$ is fixed to 1. Imagine that the association vectors can be accurately estimated when the estimated ellipses by (\ref{e42}) are located between the true inner and outer ellipses, i.e., the corresponding $\mu$ should be greater than its true value. It follows from $\eta=\beta(1-\mu^2)=1$ that $\beta$ should be large. To make this happen, we can intentionally choose a larger $\varepsilon^2$. Our simulation shows that $\varepsilon$ can be chosen from a very large range, without affecting the estimation accuracy of the association vectors.

Even after setting $\eta=1$, problem (\ref{e42}) is still a mixed integer problem and very difficult to solve. It will become more tractable if we relax the association vector $\bm{\phi}_i$ into a probability vector characterized by $0\leq \phi_{i,j}\leq1,\;\sum_{j=1}^{2}\phi_{i,j}=1,$
where $\phi_{i,j}$ is the $j$-th element of $\bm{\phi}_i$.

By doing so, problem (\ref{e42}) can be relaxed into the following form:
\begin{small}
\begin{align}\label{e44}
	&\min_{\{\bm{\phi}_i\},\;\v,\;\{\beta_i\}}\;\; \sum_{i=1}^N\; \beta_i\nonumber\\
	&\qquad{\rm s.t.}\; 0\leq \phi_{i,j}\leq1,\;\sum_{j=1}^{2} \phi_{i,j}=1,\;i=1,\ldots,N, \;\nonumber\\
	&\qquad\qquad\mbox{(\ref{e37})},\;\;\vert\v^T{\u_i}+{\bm{\phi}_i}^{T}\bm{\tau}\vert\leq\beta_i,
\end{align}
\end{small}
Problem (\ref{e44}) is a convex SOCP problem, which can be solved using some off-the-shelf softwares. Let us represent the solution of problem (\ref{e44}) as $\tilde{\bm{\phi}}_i$. An estimate of the association vector, denoted by $\hat{\bm{\phi}}_i$, can be obtained by setting the larger element of the vector $\tilde{\bm{\phi}}_i$ to 1 and the smaller element to 0.

After obtaining $\hat{\bm{\phi}}_i$, the robust ellipse fitting methods can be used to obtain the ellipse parameters. In the next subsection, we shall present a new coupled ellipse fitting method by extending the proposed MCC-VC method.
\subsection{Coupled Ellipses Fitting Based on MCC-VC}
Replacing $\bm{\phi}_i$ with $\hat{\bm{\phi}}_i$ in (\ref{e41}), the model equation becomes:
\begin{small}
\begin{align}\label{e41a}
	\hat{\bm{\phi}}^T_i{\bm{\delta}}_i=\v^T\u_i+\hat{\bm{\phi}}^T_i\bm{\tau},\; i=1,\ldots,N.
\end{align}
\end{small}
\;\; Note that the model in (\ref{e41a}) may not be ideal owing to the fact $\hat{\bm{\phi}}$ may not be equal to the true value $\bm{\phi}$.

By introducing $\tilde{\v}=[\v^T\; \eta]^T$ and
\begin{small}
\begin{align}\label{e45c}
	\tilde{\u}_{i}=\left\{\begin{array}{ll}
		{\left[\u_{i}^T\; 0\right]^{T}} & \text { if } \hat{\bm\phi}_{i}=[1,0]^{T}; \\
		{\left[\u_{i}^T\; 1\right]^{T}} & \text { if } \hat{\bm\phi}_{i}=[0,1]^{T},
	\end{array}\right.
\end{align}\end{small}
the model (\ref{e41a}) becomes a concise form: $
	\hat{\bm{\phi}}^T_i{\bm{\delta}}_i=\tilde{\v}^T\tilde{\u}_i,\; i=1,\ldots,N$. 

%

By letting $\tilde{\delta}_i=\tilde{\v}^T\tilde{\u}_i$ be the error vector and $\tilde{\v}$ be the unknown variable vector, respectively, we can similarly formulate the optimization problems based on the MCC-VC, and the resulting problems can be solved in a similar manner to the single ellipse fitting case. The parameter estimates of the two ellipses can be recovered from the solution of the MCC-VC problem as in \cite{hu2021}.

\section{Numerical Results}\label{s5}
This section verifies the robustness of the proposed MCC-VC method for single and coupled ellipses fittings using both simulated data and real images. The experiments are divided into two parts. The first part includes four subsections and mainly tests the fitting performance, in which both simulated data and real images are used to examine the performance of single ellipse fitting, data association for coupled ellipses, and coupled ellipses fitting. The second part mainly tests the ellipse detection performance, where the detection is accomplished by judging successful or failed fitting of an ellipse. In order to test the single ellipse fitting performance, we compare the proposed MCC-VC method using the Laplacian kernel (denoted by ``MCC-VC-Laplacian'') with the RANSAC method having LS as the fitting algorithm (denoted by ``RANSAC+LS'') \cite{RANSAC}, the RANSAC method having MCC-VC-Laplacian as the fitting algorithm (denoted by ``RANSAC+(MCC-VC)''), the SAREfit method (denoted by ``SAREfit'') \cite{SARE2020}, the MCC method using the Gaussian kernel (denoted by ``MCC-Gaussian'') \cite{single}, the MCC method using the Laplacian kernel (denoted by ``MCC-Laplacian'') \cite{hu2021},  the HGMM method (denoted by ``HGMM'') \cite{HGMM}, and the Szpak method (denoted by ``Szpak'') \cite{Szpak}. For coupled ellipses fitting, we first test the data association performance using the percentage rate of successful data associations, and then further compare the fitting performance of the proposed MCC-VC-Laplacian and the MCC-Laplacian \cite{hu2021} methods. The associated SOCP problems in MCC-Laplacian and MCC-VC-Laplacian are solved using the toolbox  ``ECOS'' \cite{ECOS} and SDP problem in MCC-Gaussian is solved using the Matlab toolbox ``CVX'' \cite{cvx}, where the solver is SeDuMi \cite{sedumi}.

\subsection{Single Ellipse Fitting: Simulated Data}

In the following, we generate the randomly simulated ellipses for fitting as in \cite{hu2021}. The true five ellipse parameters are set according to the following distributions: $g\sim \mathcal{U}[0,20]$, $h\sim \mathcal{U}[0,20]$, $b\sim \mathcal{U}[10,50]$, $a\sim \mathcal{U}[b+5,55]$, and $\theta\sim \mathcal{U}[-90^{\circ},90^{\circ}]$. Suppose that the noise follows the zero-mean Gaussian distribution with variance $(0.005b)^2$. The normalized root mean square error (NRMSE) is used to evaluate the fitting performance, which is defined by $
	{\rm NRMSE} = \sqrt{\frac{1}{KM}\sum_{k=1}^K\sum_{m=1}^{M}\|\hat{\q}_{mk}-\q_{k}\|^2}$,
where $K$ and $M$ are the numbers of the generated ellipses and the Monte Carlo (MC) runs for each ellipse, and $\hat{\q}_{mk}$ and $\q_{k}$ represent the estimated and true parameters of the $k$th ellipse in the $m$th MC run. In the following, we set $K=100$ and $M=500$ to compute the NRMSE.
Due to the existence of outliers, the fitting possibly fails in a few runs. To obtain a meaningful NRMSE, it is computed by discarding the results of these runs. Several conditions are used to identify the failed runs. Here, we adopt the same rule with that in \cite{hu2021} to identify a failed fitting.

\subsubsection{Scenario 1: Uniformly Distributed Outliers}
\begin{figure*}[htbp]  
		\subfigure[]{  
			\includegraphics[height=3.4cm,width=4.2cm]{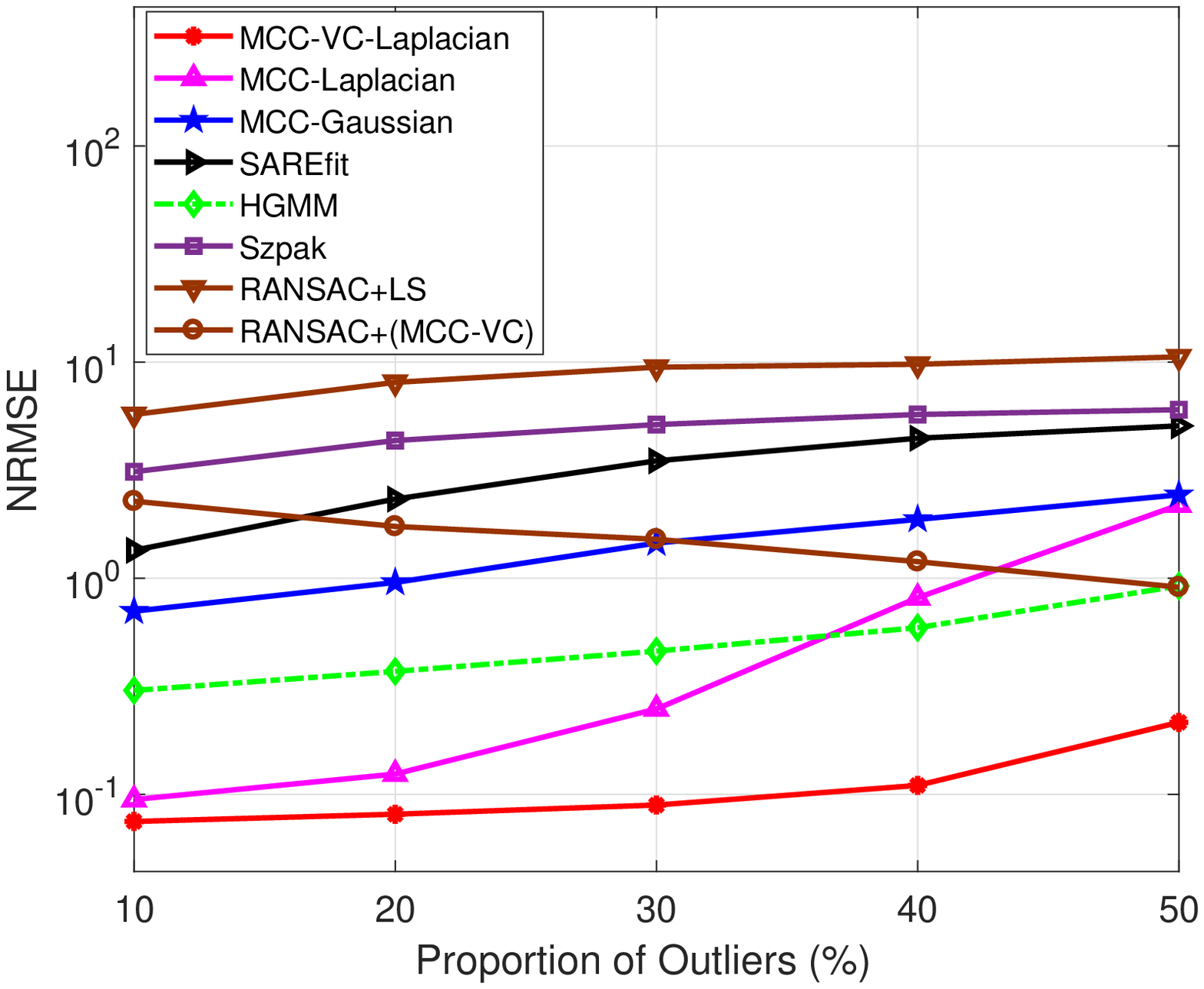}  }  
		\subfigure[]{  
			\includegraphics[height=3.4cm,width=4.2cm]{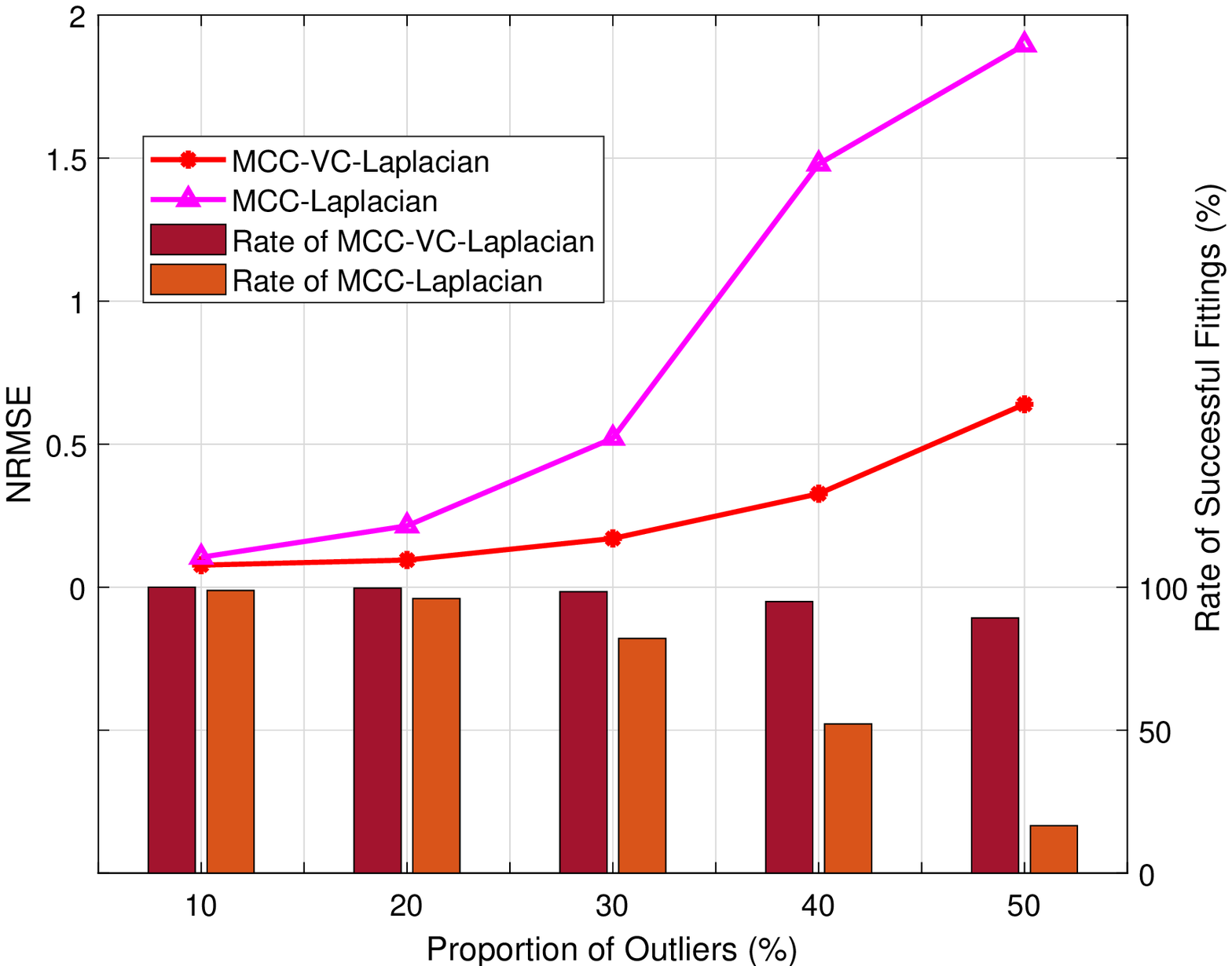}  } 
		\subfigure[]{  
        		\includegraphics[height=3.4cm,width=4.2cm]{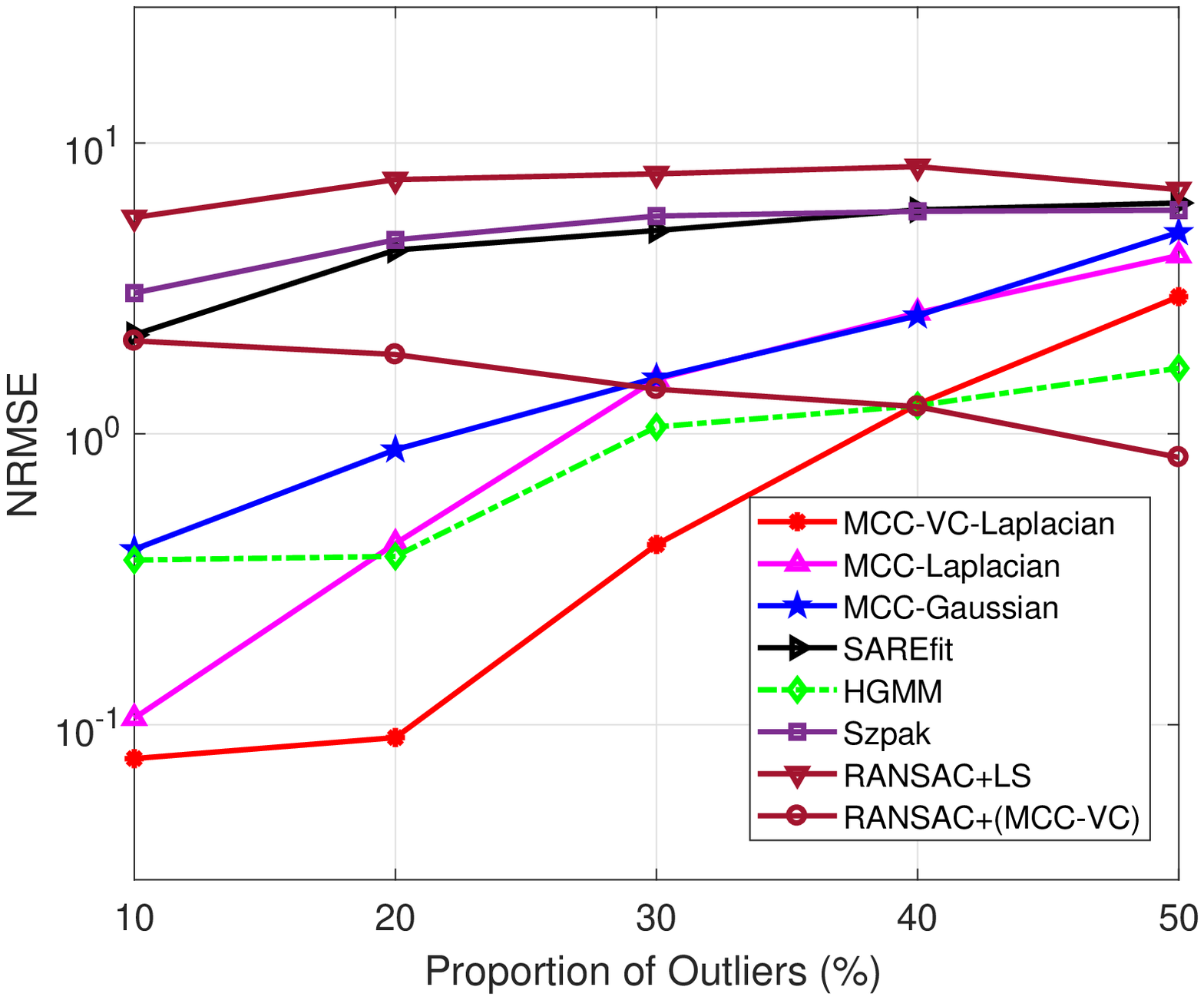}  }  
       	\subfigure[]{  
        		\includegraphics[height=3.4cm,width=4.2cm]{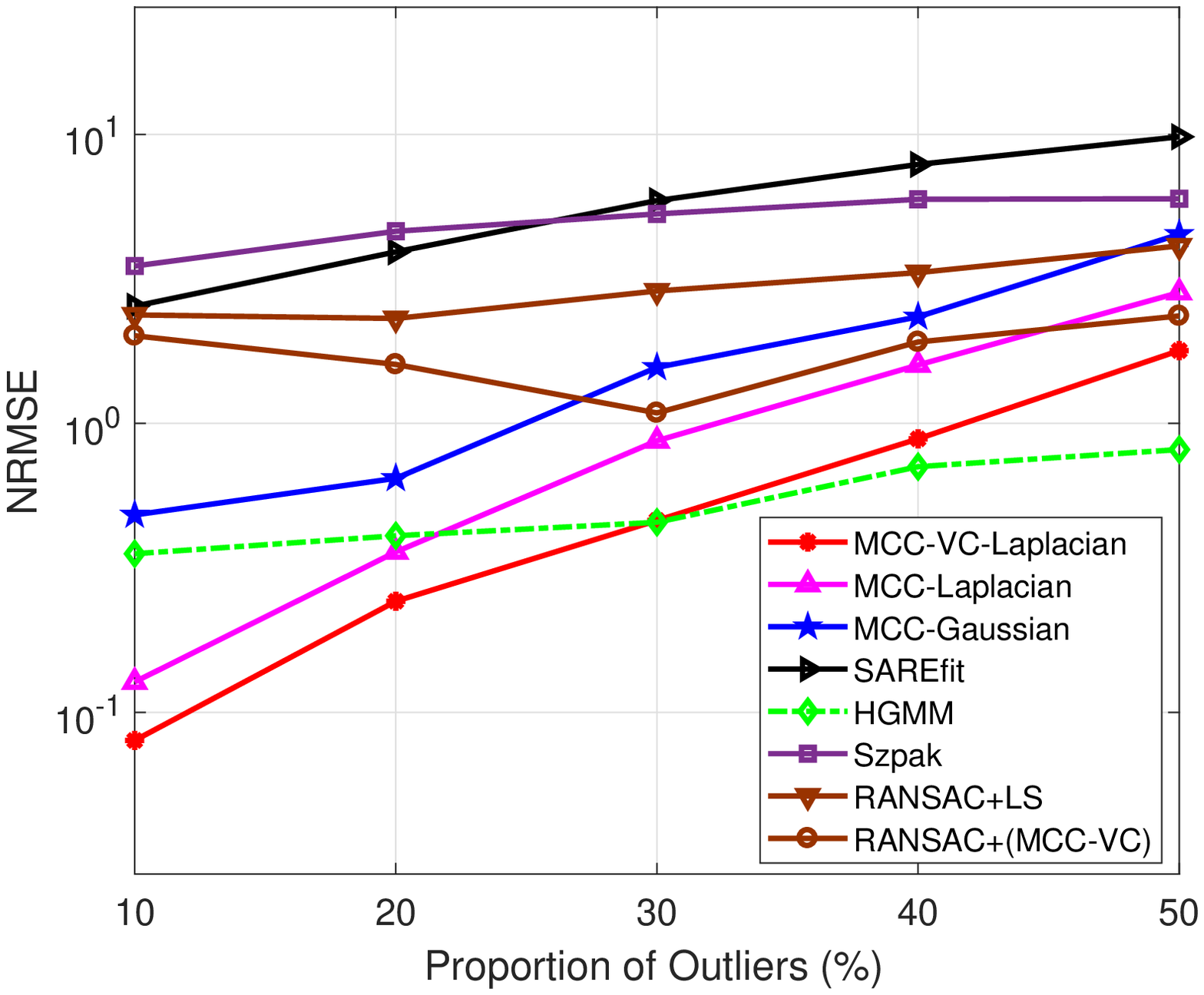}  }  		
     	\caption{NRMSEs of all compared methods in different distributed outliers scenarios. (a) uniformly distributed $\mathcal{U}(-b,b)$; (b) NRMSE and rate of successful fittings for MCC-VC and MCC methods in uniformly distributed $\mathcal{U}(-b,a)$; (c) cluster-like distributed; (d) One-sidedly distributed.} \label{f1}  
\end{figure*}

\begin{figure*}[htbp]  
	\begin{center}
		\subfigure[]{  
			\includegraphics[height=3.4cm,width=4.2cm]{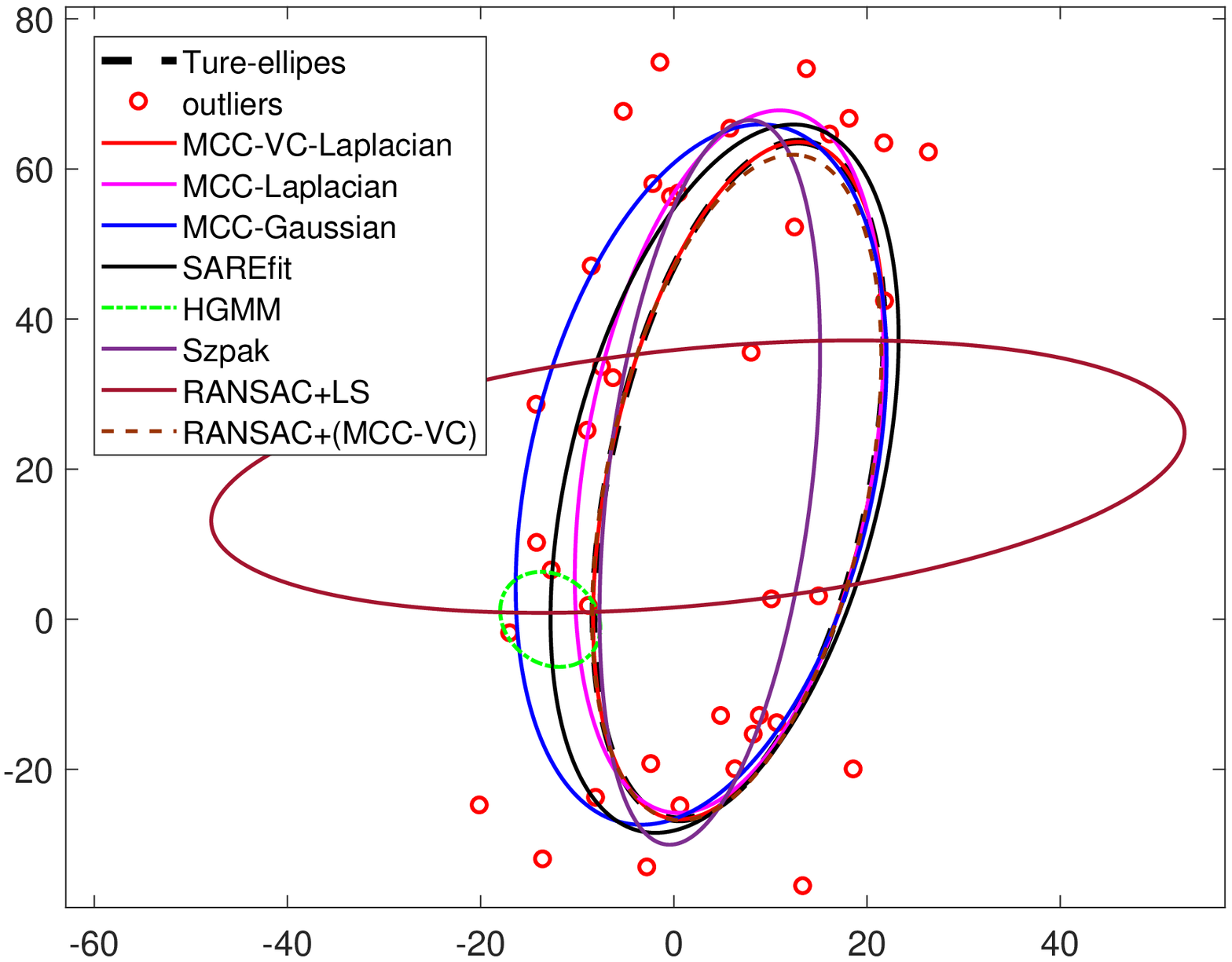}  } 
 		\subfigure[]{  
 			\includegraphics[height=3.4cm,width=4.2cm]{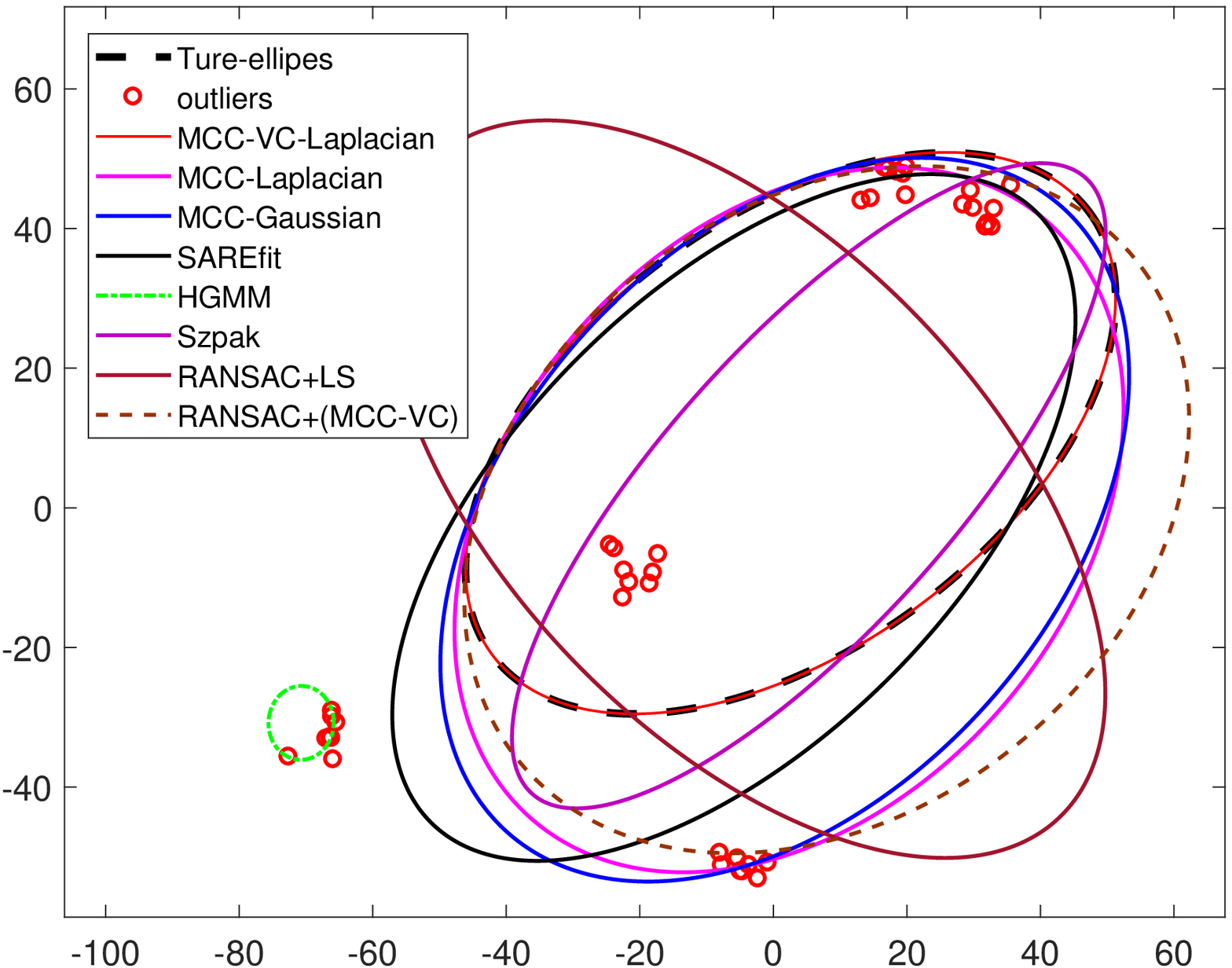} } 			  
		\subfigure[]{  
			\includegraphics[height=3.4cm,width=4.2cm]{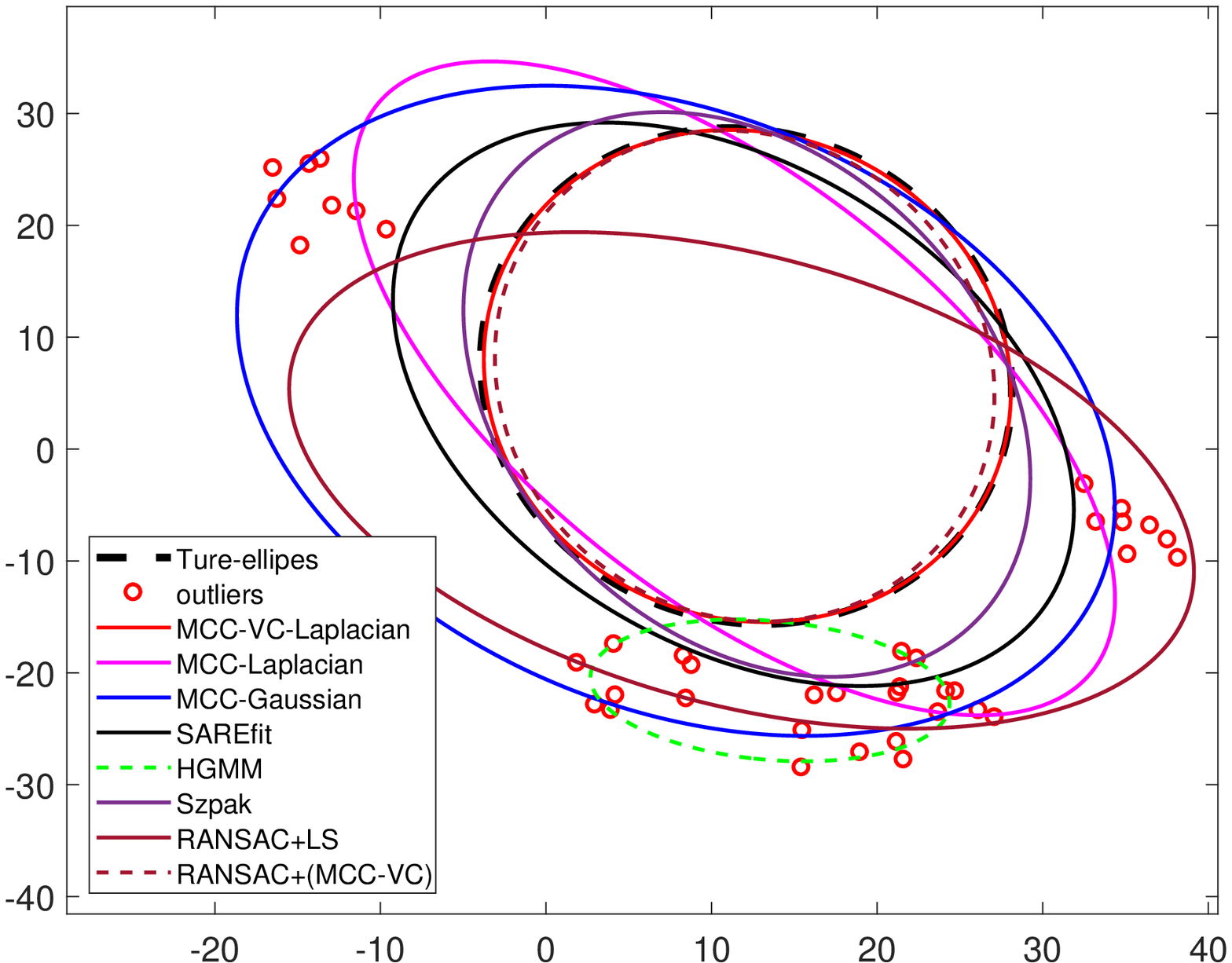}  } 		
		\subfigure[]{  
			\includegraphics[height=3.4cm,width=4.2cm]{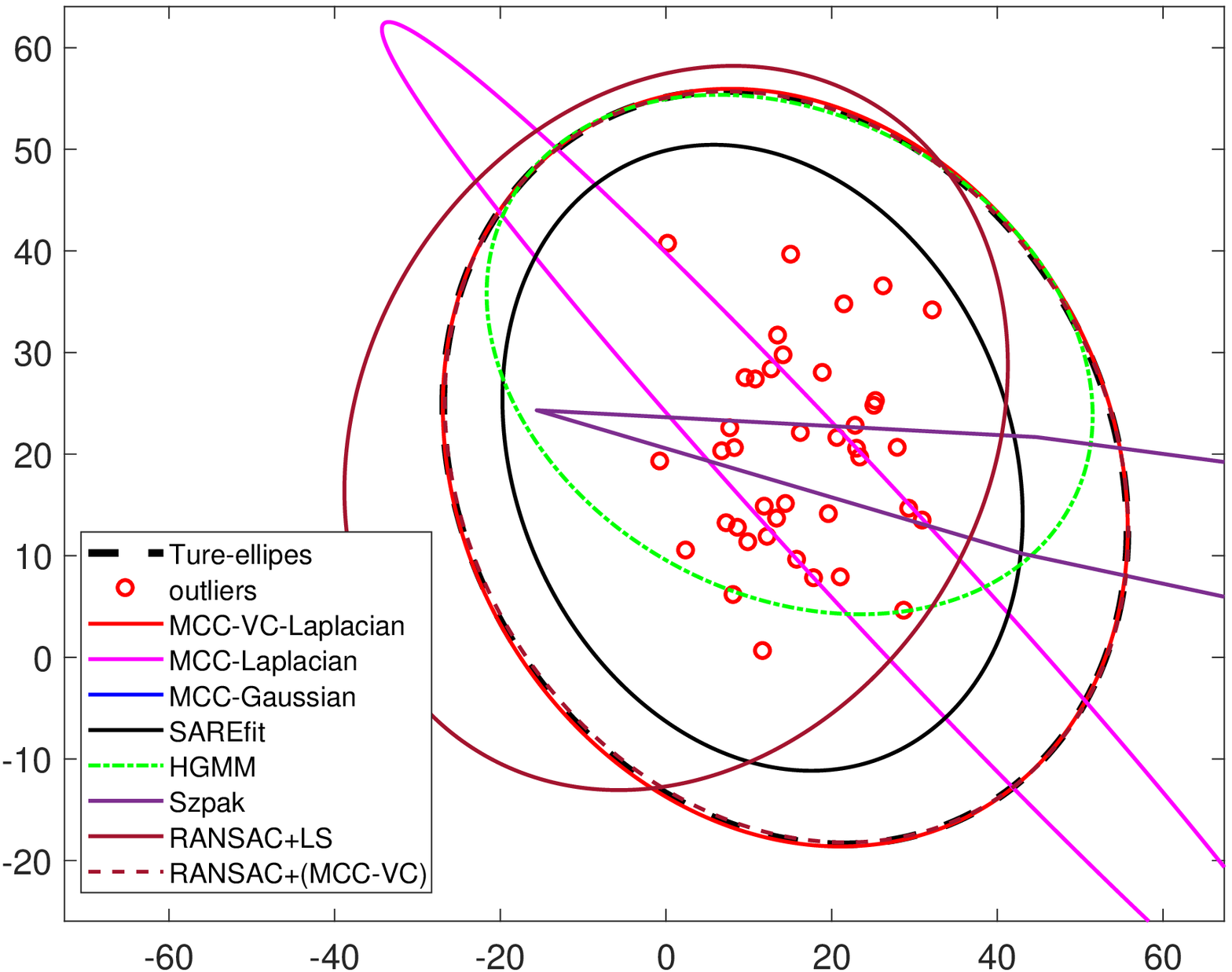}  }  		
		\caption{Illustration of an incorrect fitting case of the compared methods for single ellipse fitting having $40\%$ outliers in different distributed outliers scenarios. (a) uniformly distributed outliers; (b) cluster-like outliers: both outside and inside of the ellipse; (c) cluster-like outliers: outside of the ellipse; (d) cluster-like outliers: inside of the ellipse.} \label{f2} 
	\end{center}  
\end{figure*}

In this scenario, the total number of data points is $N=100$, and the proportion of outliers varies from 10\% to 50\%. The values $v_i^x$ and $v_i^y$ for the outliers are uniformly generated according to $\mathcal{U}(-b,b)$ or $\mathcal{U}(-b,a)$ for the simulated scenarios where the outlier distribution is either zero-mean or non-zero-mean. Fig. \ref{f1}(a) shows the NRMSE performance as the proportion of outliers increases. Generally speaking, the proposed MCC-VC-Laplacian method performs better than the other methods, including the MCC-Laplacian method, especially when the proportion of outliers increases from $30\%$ to 50\%. Note that the outliers in this scenario have zero mean, implying that MCC-VC-Laplacian is still able to work well even for zero-mean outliers. Correspondingly, Table \ref{t1} shows the percentage rates of successful fittings for different methods. In this simulation scenario, the proposed MCC-VC-Laplacian almost always successfully fits the ellipses even when the proportion of outliers is 50\%. By contrast, the other methods may fail, especially when the proportion of outliers is large. As an illustration, Fig. \ref{f2}(a) shows the ellipses fitted by the eight methods in a typical MC run. It can be seen that the ellipse generated by MCC-VC-Laplacian fits the true ellipse very well, and better than the other methods. Interestingly, after replacing the fitting algorithm in the RANSAC method with MCC-VC-Laplacian (RANSAC+(MCC-VC) in Fig. \ref{f1}(a) and Fig. \ref{f2}(a) and Table \ref{t1}), its performance is significantly improved compared with the original RANSAC method (RANSAC+LS). This further indicated that the good performance of the proposed method when dealing with the fitting on a small data set. Same observations can be found in the following experiments. The results of Fig. \ref{f1}(b) confirm that the MCC-VC-Laplacian method does have great advantages in the case with non-zero-mean distribution of outliers, in terms of both the fitting accuracy and successful fitting rate.

\begin{table}[htb]
	\centering
	\caption{Rate (\%) of Successful Fittings for Single Ellipse  (50000 MC Runs in Total): Uniformly Distributed Outliers}\label{t1}
	\begin{tabular}{c|c|c|c|c|c}\hline\hline
		\diagbox{Method}{Prop.(\%)}&10&20&30&40&50\\\hline
		MCC-VC-Laplacian  &100&100&100&100&99.96\\\hline
		MCC-Laplacian &100&100&98.82&87.41&54.79\\\hline
		MCC-Gaussian &89.23&74.86&51.56&27.63&15.11\\\hline
		SAREfit &96.53&91.00&81.91&69.52&59.67\\\hline
		HGMM &24.15&16.10&9.43&5.16&2.91\\\hline
		Szpak &85.41&64.29&40.29&20.62&9.07\\\hline
		RANSAC+LS &14.67&10.82&5.08&1.48&0.38\\\hline				
		RANSAC+(MCC-VC) &86.93& 94.10& 94.74&91.28&86.95\\\hline\hline
	\end{tabular}
\end{table}


\subsubsection{Scenario 2: Cluster-Like Outliers}
In this scenario, we simulate a typical outlier distribution, where cluster-like outliers are randomly distributed around the true ellipse. Specifically, we generate five clusters for each ellipse.  The cluster center for each cluster is first generated similarly to the outliers in Scenario 1, and the other points of this cluster are then randomly generated in the $15 \times 15$ square area located at the cluster center. Each cluster has the same number of outliers. The same as in Scenario 1, we vary the proportion of outliers from 10\% to 50\%. The fitting NRMSEs are shown in 
Fig. \ref{f1}(c) and the rates of successful fittings are given in Table \ref{t2}. Although HGMM has better fitting performance than MCC-VC-Laplacian in NRMSE when the proportion of outliers is 50\%, its successful fitting rate is far from satisfactory. In general, the proposed MCC-VC-Laplacian still significantly outperforms the other methods. Comparison between the results for the uniformly distributed outliers and the cluster-like outliers reveals that the cluster-like outliers will result in larger fitting errors and higher rate of failed fittings, indicating that the ellipse is more difficult to fit in this scenario. Fig. \ref{f2}(b) illustrates the fitting results in a typical MC run, which also shows the distribution of the outliers. It clearly indicates the better fitting performance of the proposed method. 
\begin{table}[htb]
	\centering
	\caption{Rate (\%) of Successful Fittings for Single Ellipse  (50000 MC Runs in Total): Cluster-Like Outliers}\label{t2}
	\begin{tabular}{c|c|c|c|c|c}\hline\hline
		\diagbox{Method}{Prop.(\%)}&10&20&30&40&50\\\hline
		MCC-VC-Laplacian  &100&100&98.96&77.42&53.18\\\hline
		MCC-Laplacian &100&95.21&63.55&40.09&27.57\\\hline
		MCC-Gaussian & 90.40&63.31&30.55&20.32&9.26\\\hline
		SAREfit &95.62&72.32&38.22&30.82&31.57\\\hline
		HGMM &26.54&17.12&11.16& 4.17&1.73\\\hline
		Szpak &85.11&70.05&50.05&38.94&20.48\\\hline
		RANSAC+LS &14.68&11.47&3.71&7.10&3.15\\\hline				
		RANSAC+(MCC-VC) &87.22&93.21&92.50&86.34&70.39\\\hline\hline
	\end{tabular}
\end{table}
As a special case of the cluster-like distribution, the one-sided distribution of outliers have non-zero mean of error samples. This case is very common in real images as shown later. To test the superior performance of the proposed MCC-VC-Laplacian method in dealing with such a case, we design the following experiment. All outliers are similarly divided into several clusters, but they are located on one side of the ellipse. To confine the outliers inside the ellipse, we set the distance between the cluster centers and the ellipse center to a small range $0.25b-0.5b$. The outliers outside the ellipse are generated by setting the distance between the cluster centers and the ellipse center to a large range $1.5b-2b$. Fig. \ref{f1}(d) shows the NRMSE curves and Table \ref{t3} gives the rates of successful fittings as the proportion of outliers varies from 10\% to 50\%. As shown in Fig. \ref{f1}(d), the fitting errors increase significantly as compared to those in Scenario 1, indicating that the ellipse with one-sided outliers is more difficult to fit well. However, the proposed MCC-VC-Laplacian method performs much better than the other methods, in terms of both NRMSE and successful fitting rate. Similarly, Fig. \ref{f2}(c) and Fig. \ref{f2}(d) respectively confirm the fitting results of MCC-VC-Laplacian are better than the other methods when the outliers are located inside and outside the ellipse in two typical MC runs.

\begin{table}[htb]
	\centering
	\caption{Rate (\%) of Successful Fittings for Single Ellipse (50000 MC Runs in Total): One-sidedly Distributed Outliers}\label{t3}
	\begin{tabular}{c|c|c|c|c|c}\hline\hline
		\diagbox{Method}{Prop.(\%)}&10&20&30&40&50\\\hline
		MCC-VC-Laplacian  &99.99&97.97&83.20&68.08&46.13\\\hline
		MCC-Laplacian &97.83&84.37&76.15&46.29&24.49\\\hline
		MCC-Gaussian &81.19&61.18&43.88&37.01&32.40\\\hline
		SAREfit &81.60&70.59&50.92&46.13&42.54\\\hline
		HGMM &23.75&18.48&15.16&9.43&6.12\\\hline
		Szpak &49.38&29.29&21.52&12.96&7.18\\\hline
		RANSAC+LS &9.62&8.72&7.26&6.79&6.57\\\hline				
		RANSAC+(MCC-VC) &91.36&84.21&72.15&63.34&51.34\\\hline\hline
	\end{tabular}
\end{table}



\subsection{Coupled Ellipses Fitting: Simulated Data}

In the coupled ellipses fitting, the ellipse parameters $\{g, h, a, b, \theta\}$ are generated in the same way as the single ellipse fitting case. The proportional parameter $\mu$ is generated randomly according to the uniform distributions $\mu\sim \mathcal{U}(0,1)$.
100 data points are collected for each ellipse (and hence, the number of total data points is $N=200$), possibly with outliers included. The normal data points are generated by uniformly sampling over the ellipses, and the outliers are simulated according to the uniform distribution $\mathcal{U}(-b,b)$. The associations between the data points and the coupled ellipses are unknown before the fitting.

Due to the possible incorrect association of data points in Step 1 and the existence of outliers, we take the same measures as mentioned before to discard the failed fitting results to obtain meaningful NRMSE.

\subsubsection{Effect of $\varepsilon$ on Association}

\begin{figure}[htb]
	\centering
	\includegraphics[height=3.5cm,width=5cm]{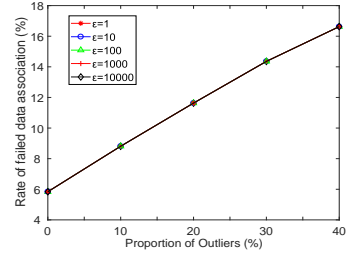}
	\caption{Sensitivity of the value of $\varepsilon$ on the association accuracy.} \label{f3}
\end{figure}

As aforementioned, $\varepsilon$ cannot be arbitrarily chosen owing by fixing $\eta$ to 1 during the data association step. In this experiment, we study the effect of choosing different values of $\varepsilon$ on data association. The percentage of incorrect associations of both normal and outlier data points is examined, as the proportion of outliers increases from 0 to 40\%, when $\varepsilon$ takes the values of 1, 10, 100, 1000 or 10000. The results are shown in Fig. \ref{f3}, which indicates that the choice of $\varepsilon$ has nearly no effect on the association vector estimation, although the ellipse parameter vector may not be accurately estimated. In the following, $\varepsilon$ is set to 1.

%

\subsubsection{Performance of Coupled Ellipses Fitting by Varying the proportion of Outliers}

In this experiment, we consider the scenario that outliers exist inherent in the data points. The noise STD is fixed at $0.005b$, and the proportion of outliers varies from 0 to 40\%. The results are given in Table \ref{t5}.
The results remain encouraging. Let us take the last column as an example. With the incorrectly associated points included, the proportion of the outliers is greater than 40\% under this setting. Even in this challenged case, the rate of successful fitting is still greater than 97.7\%, implying that the proposed data association method is quite effective for use in practice in a harsh environment.
\begin{table}[htb]
	\centering
	\caption{Performance of Coupled Ellipses Fitting by Varying the Proportion of Outliers}\label{t5}
	\begin{tabular}{c|c|c|c|c}\hline\hline
		\diagbox{Performance}{Prop.(\%)}&10&20&30&40\\\hline
		Incorrect association rate (\%)&8.8&11.6&14.4&16.6\\\hline
		Successful fitting rate-MCC (\%)&98.5&92.8&82.2&71.3\\\hline
		Successful fitting rate-MCC-VC (\%)&100&99.8&99.7&97.7\\\hline
		NRMSE-MCC &0.71&1.26&1.55&1.74\\\hline
		NRMSE-MCC-VC &0.09&0.16&0.57&1.07\\\hline\hline
	\end{tabular}
\end{table}
\subsection{Single Ellipse Fitting: Real Data}
In this subsection, we apply five methods to fit the ellipses in real images, including the Voyager aircraft \cite{Voyager}, Mars, and globe images. Before the fitting, the data points of these images are extracted through a series of preprocessing steps including the image segmentation, the morphological operations, and the edge detection techniques. It is not difficult to imagine there are a large number of outliers in the extracted data points, and the outliers do not necessarily follow a zero-mean distribution. 
\subsubsection{Voyager Aircraft Image}
\begin{figure}[htb]
	\centering
	\includegraphics[height=5.3cm,width=9.1cm]{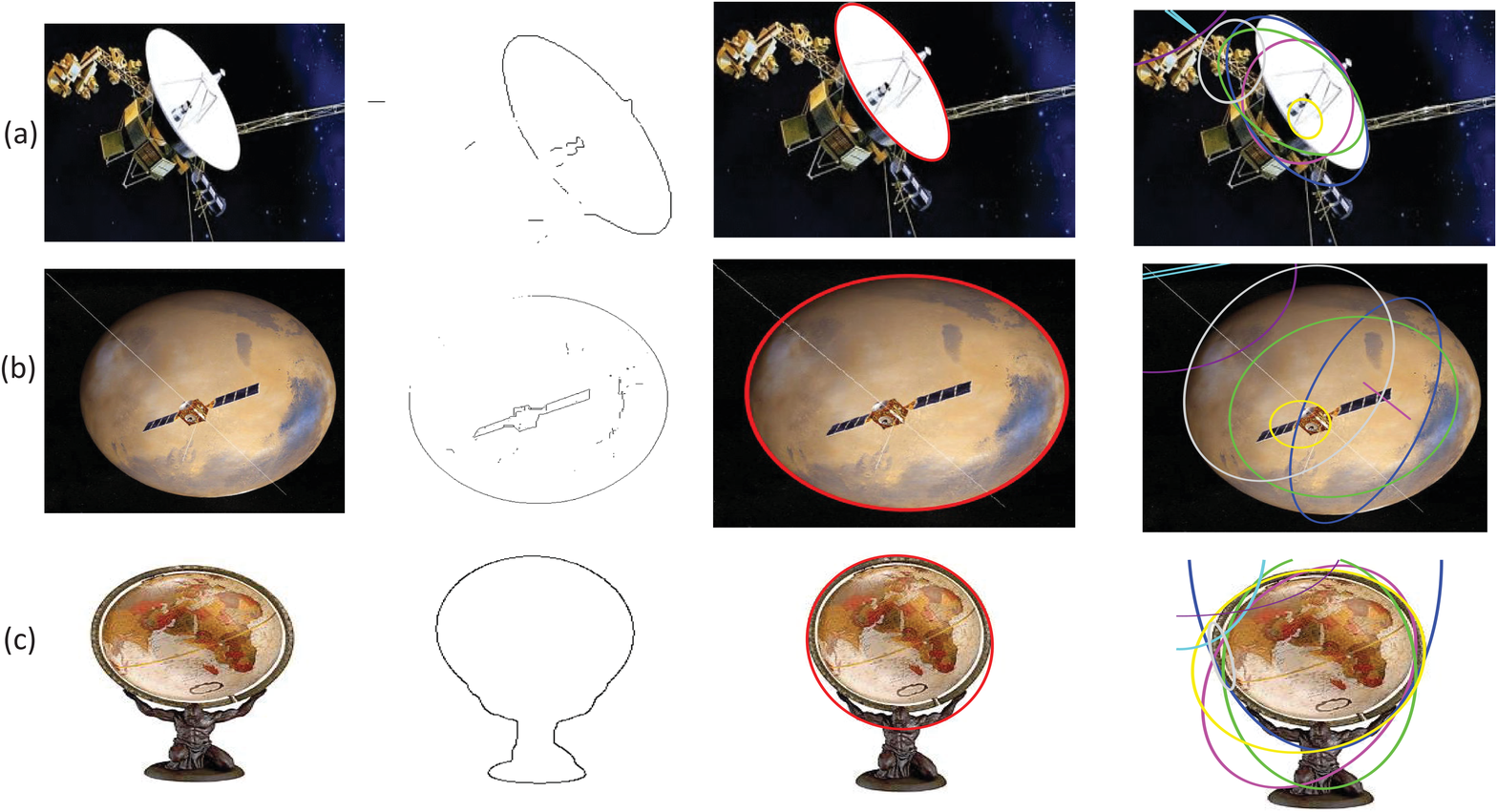}  
	\includegraphics[height=1cm,width=9cm]{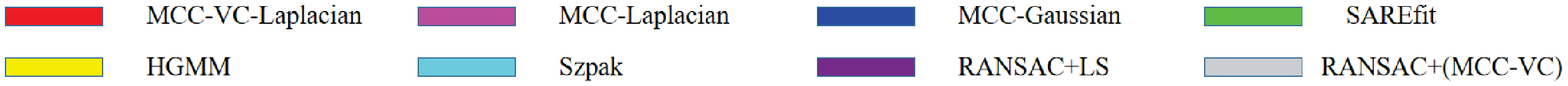}
	\caption{Left to right columns: input images, data points, fitting results attained by our method, and fitting results attained by all compared methods: row (a): Voyager aircraft image; row (b): mars image; row (c): globe image. } \label{f4}
\end{figure}

The Voyager aircraft image fitting process and results are shown in  row (a) of Fig. \ref{f4}, in which the proportion of the outliers\footnote{The outliers are recognized in the following way. First, the true ellipse parameters are obtained through a manual measurement tool. The points with errors greater than 0.1 in all data points are regarded as outliers.} is about 21.47\%. The proportion is larger than that in \cite{hu2021}, which is generated by setting the Sobel operator parameter to a smaller value of 0.17 as compared to 0.2 in \cite{hu2021}. The fitting results of the MCC-VC-Laplacian method and other methods are shown in the third column and fourth column of row (a), respectively. Obviously, MCC-Laplacian \cite{hu2021} fails to fit the ellipse but MCC-VC-Laplacian is still successful when the number of the outliers is larger, indicating that the proposed method is more robust to larger amount of  outliers.

\subsubsection{Mars Image}
The Mars image is selected from the Caltech 256 Dataset \cite{256dataset}, labeled as $137\_0008$. The fitting process and results are shown in row (b) of Fig. \ref{f4}. Similarly, the parameter of the Sobel operator is set to 0.042 for this image to include more outliers as compared to \cite{hu2021}. The proportion of the outliers is about 54.51\% in this image. The fitting results of MCC-VC-Laplacian and the other methods are shown in the  third column and fourth column, of row(b), respectively, we see that the proposed method successfully fits the ellipse but the others fail.

\subsubsection{Globe Image}
The globe image is also selected from \cite{256dataset}, labeled as $053\_0080$. The fitting process and results are shown in row (c) of Fig. \ref{f4}. Different from the previous two images, the Canny detector is used for edge extraction of this image, where the correlation coefficient is set to 0.5. The outliers in the extracted data points form a one-sided distribution, due to the base holding the globe. The proportion of the outliers is about 33.27\%. For this image, the proposed MCC-VC-Laplacian method has the best fitting performance compared to the other methods, as shown in the third column and fourth column of row (c), confirming the robustness of the proposed method to one-sided outliers.

\subsection{Coupled Ellipses Fitting: Real Data}

\begin{figure}[htbp]  
		\begin{center}  	  
	\subfigure[]{  
		\includegraphics[height=2cm,width=2.5cm]{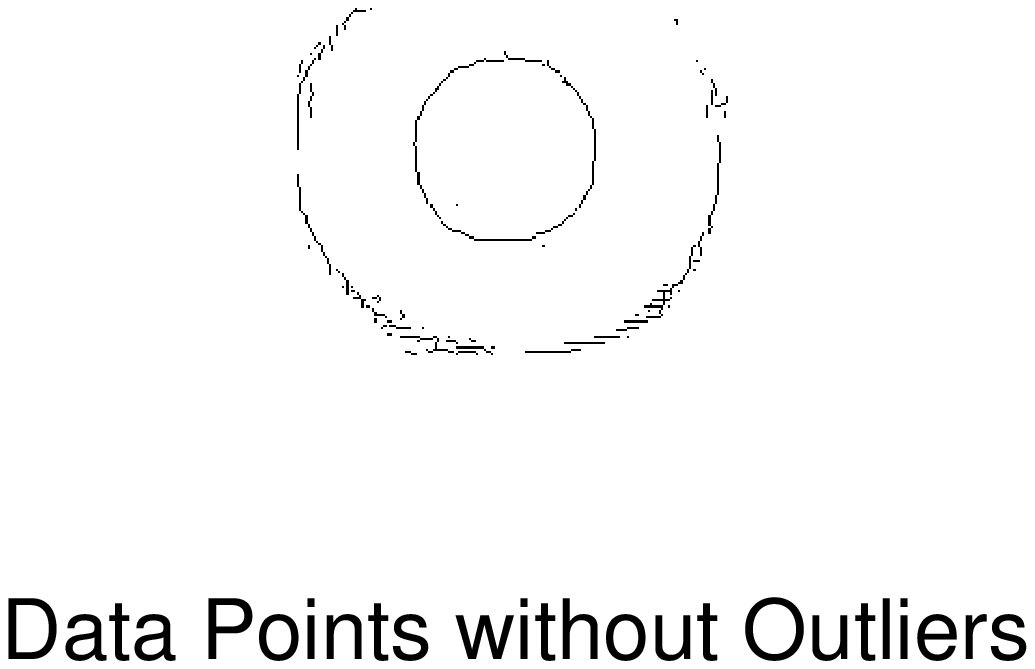}  } 	
	\subfigure[]{  
		\includegraphics[height=2cm,width=2.5cm]{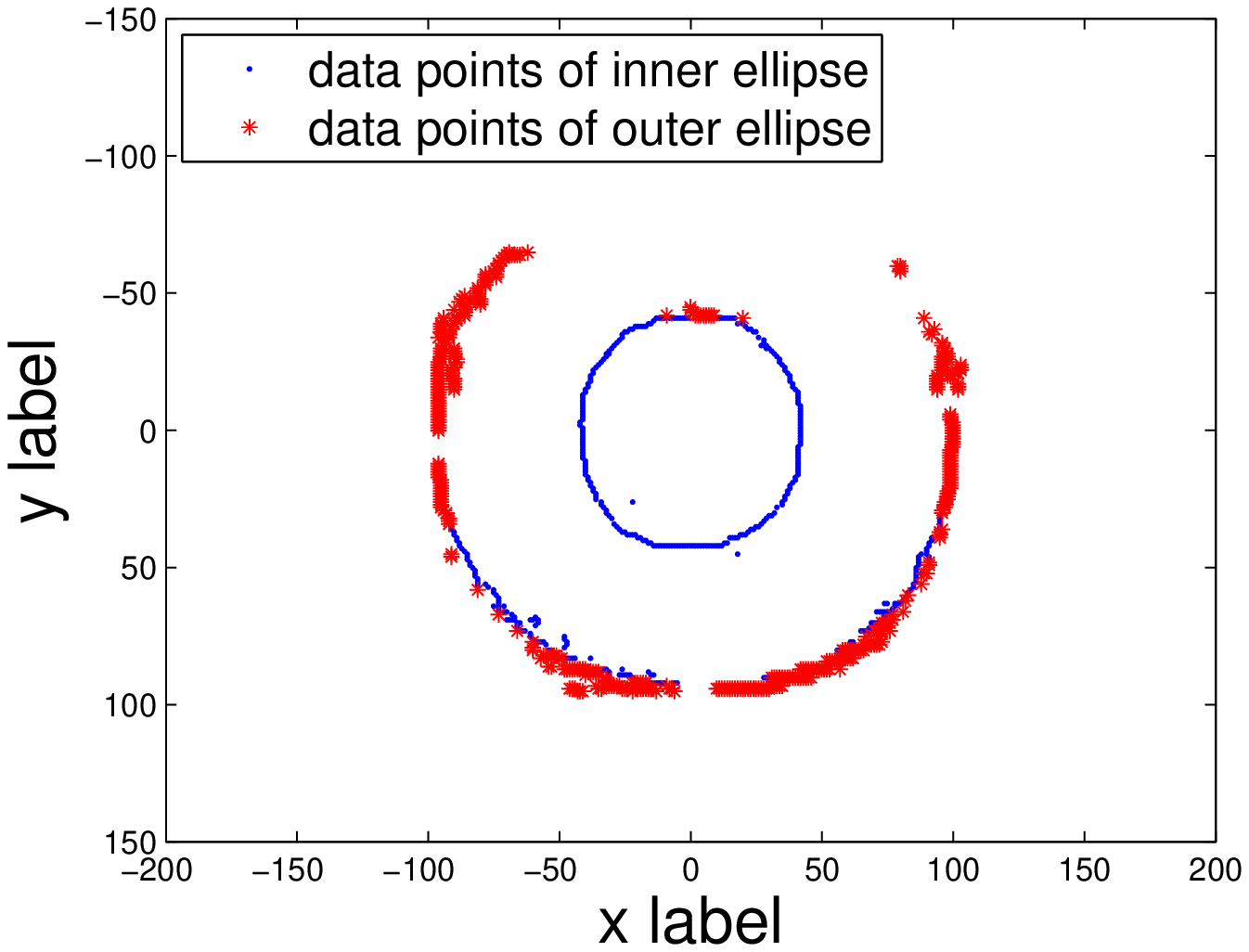}  }
	\subfigure[]{  
		\includegraphics[height=2cm,width=2.5cm]{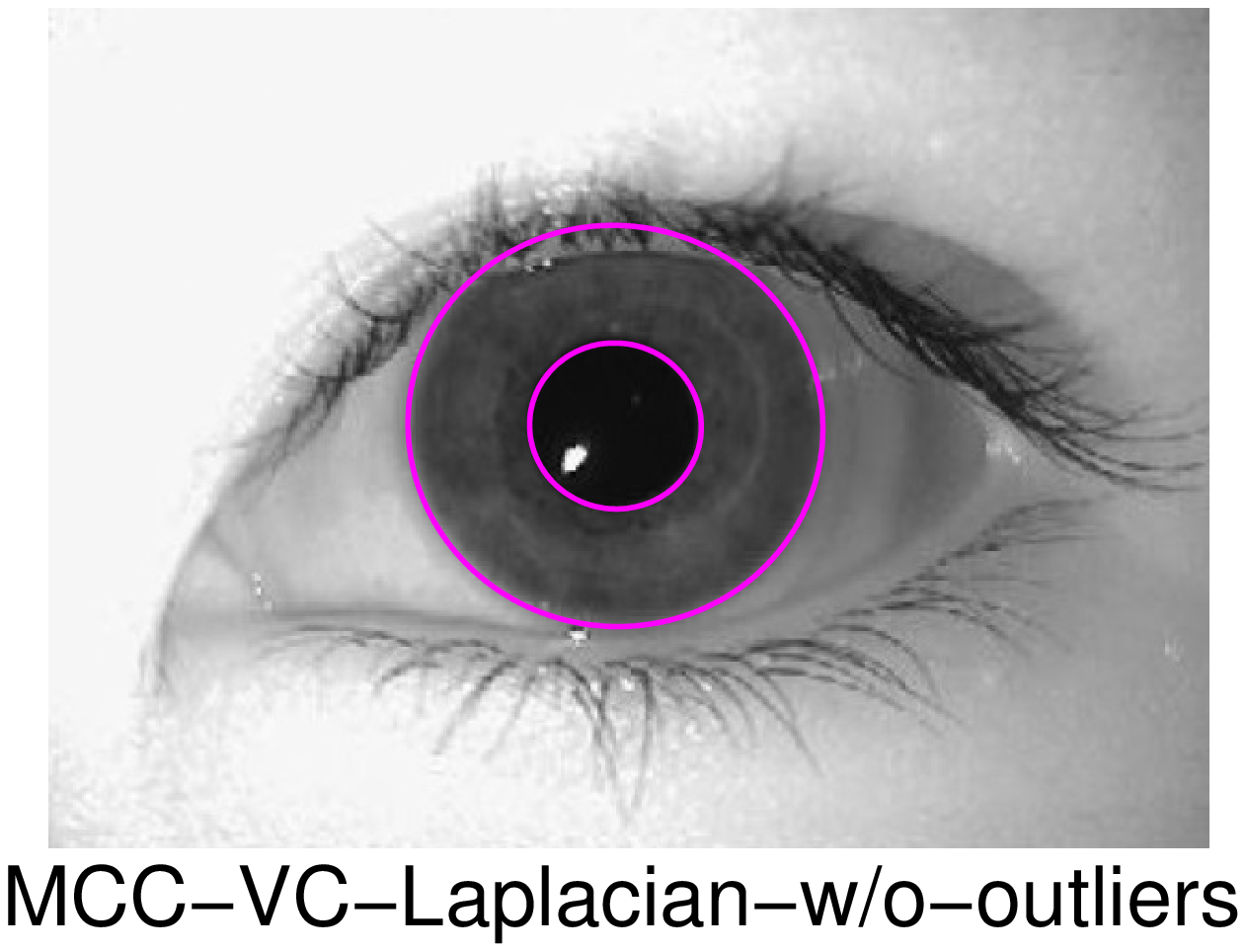}  } 		
	\subfigure[]{  
		\includegraphics[height=2cm,width=2.5cm]{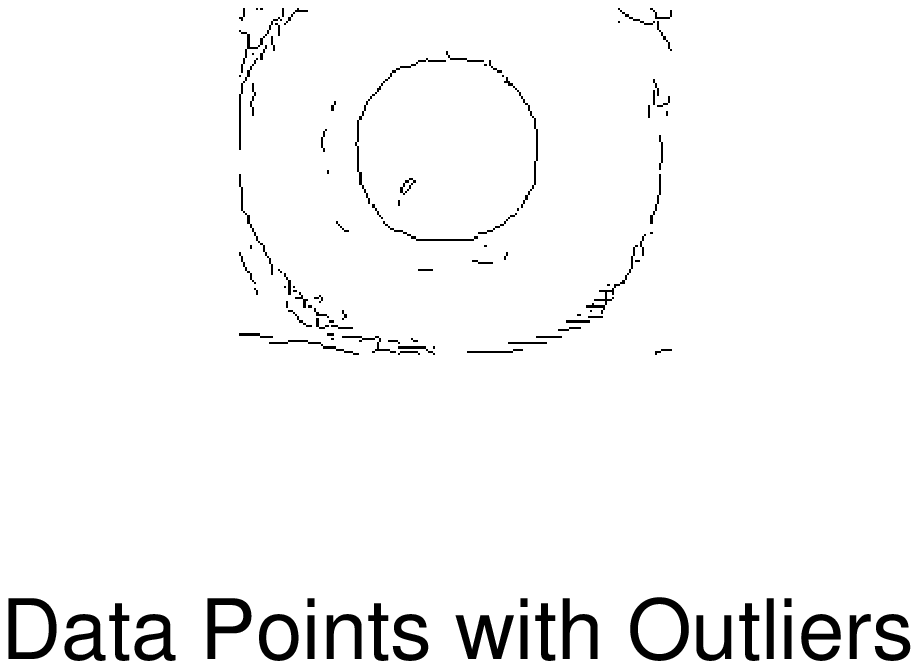}  } 		
	\subfigure[]{  
		\includegraphics[height=2cm,width=2.5cm]{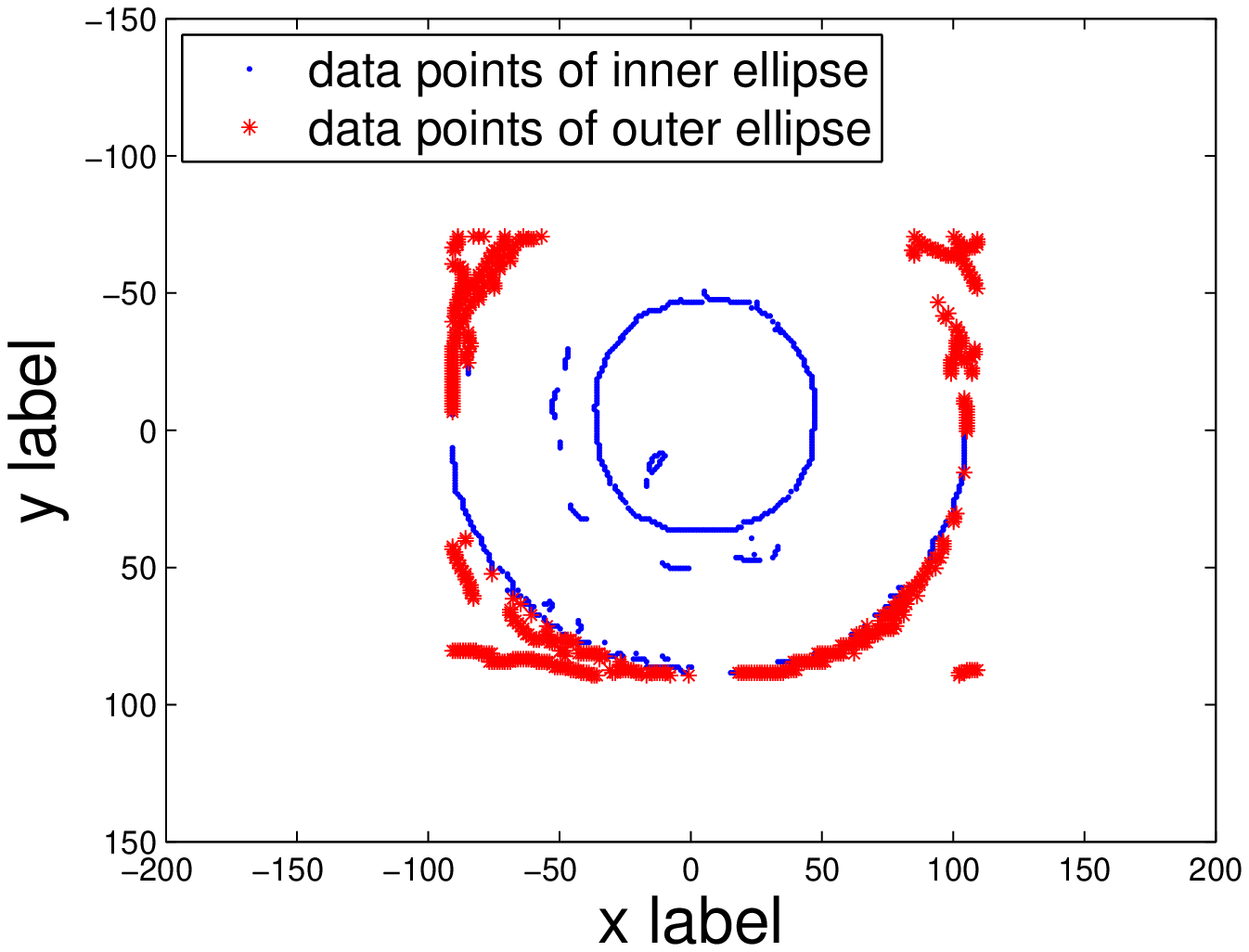}  }
	\subfigure[]{  
		\includegraphics[height=2cm,width=2.5cm]{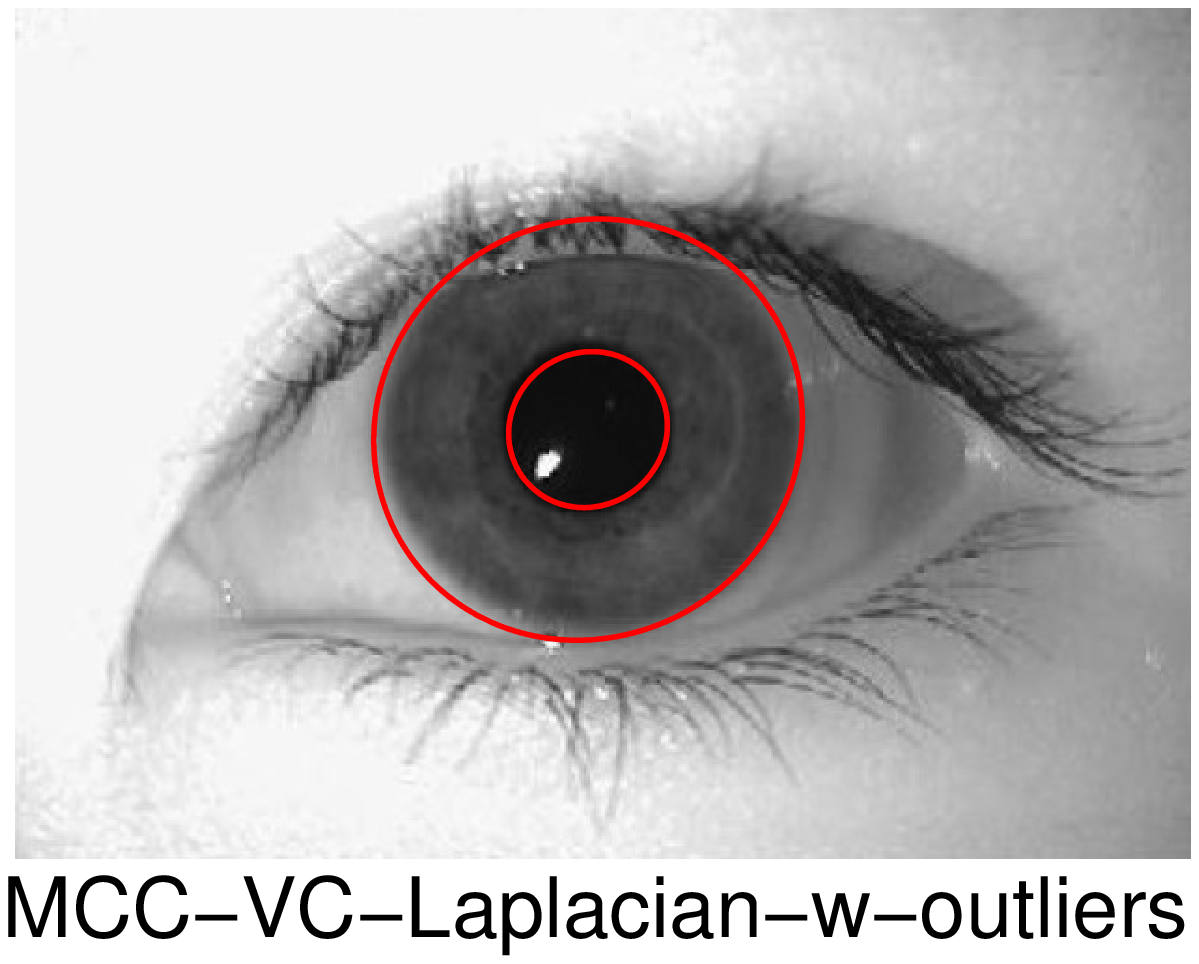} } 	 		
	\caption{Coupled ellipses fitting for iris image, scenario without outliers (first row), scenario with outliers (second row). Left to right columns: data points, association results attained by our method, and fitting results attained by our method. } \label{f5} 
		\end{center}  
\end{figure}

In this subsection, the proposed method is applied to fit the coupled ellipses in an iris image. To demonstrate the robustness of the proposed method, two scenarios without and with outliers in the data points are investigated. The data points extracted from iris image in both scenarios are shown in Fig. \ref{f5} (a)(d). In the case of having outliers, the outliers account for 24.98\% of the total data points.  The results of the data association and the fitting are illustrated in Fig. \ref{f5} (b)(e) and Fig. \ref{f5} (c)(f), respectively. In Fig. \ref{f5} (b)(e) the blue dots and and the red stars represent the data points associated with the inner and outer ellipses, respectively. Clearly, there exist some incorrectly associated points, and they are regarded as outliers, and hence, the percentage of the outliers\footnote{The outliers are recognized by comparing the data points of Fig. \ref{f5}  (b)(e).} is greater than 24.98\% when doing the fitting. However, the fitting is still successful, indicating the incorrectly associated data points are handled without difficulty by the fitting method in Step 2.

\subsection{Multiple Ellipses Fitting}

In this subsection, we show that the proposed method has the ability of accurately fitting multiple ellipses, with the help of ellipse detection. It is well known that the popular ellipse detection methods \cite{Prasad2012,Lu2020,AAMED,TCYBcell} involve ellipse fitting using sampled data points. Specifically, the principle of many detection methods is to detect arcs by the position relationship among pixel points. After grouping or deleting the arcs through basic fitting methods such as the direct least squares method \cite{directa}, the candidate ellipses are formed, and then the qualified ellipses are selected from the candidates through screening. The detection methods focus on the detection performance, and their fitting performance may not be satisfactory. Our aim is to first apply the ellipse detection method to find the ellipses in a given image, and then further improve the fitting performance based on the coarse position and size information of the detected ellipses. More specifically, the idea is to divide the data points into groups, each corresponding to one detected ellipse. To this end, we first form the equations of the detected ellipses using the parameters of the detected ellipses obtained from the detection methods.  We then apply the coordinates of each data point to the equations and compute the errors of each data point relative to all ellipse equations. For each ellipse, we can extract the data points with relatively small errors by setting a threshold. By doing so, the extracted points belong to the ellipse with a high probability. This process can be regarded as association, i.e., associate the data points to particular ellipses. Apparently, outliers may be introduced in the association process, and the value of the threshold determines the number of outliers. The smaller the threshold, the less the number of outliers, which, however, may filter out useful points.   

\begin{figure}[htb]
	\includegraphics[height=2cm,width=9cm]{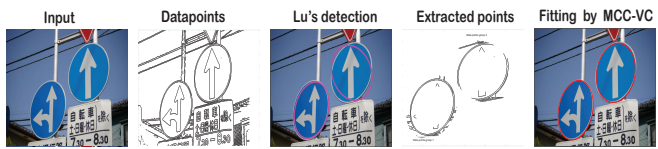}
	\caption{Example for multiple ellipses fitting after detection.} \label{f6}
\end{figure}

\begin{figure}[htb]
	\centering
	\includegraphics[height=8.5cm,width=9.5cm]{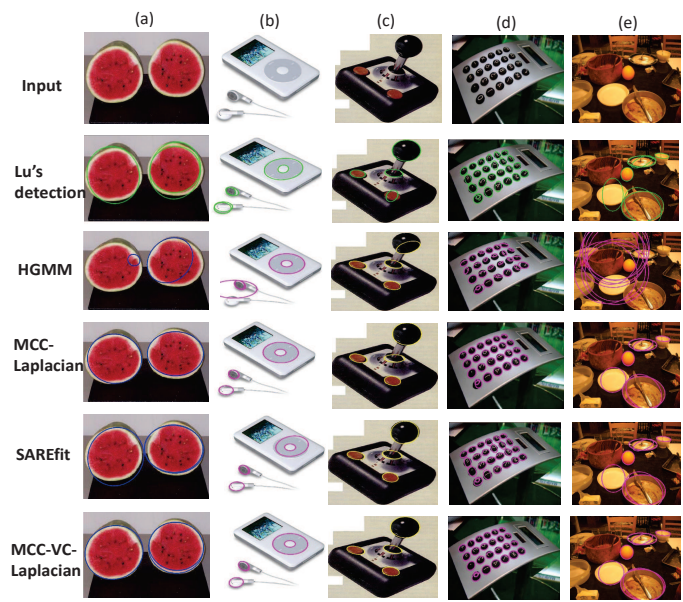}
	\caption{Fitting performance comparison for multiple ellipses fitting after ellipse detection. (1) The first row of images are the original images from the Caltech 256 Dataset \cite{256dataset}; (2) The second row shows the results of Lu's detection method; (3) The third, fourth, and fifth rows show the fitting results obtained by the existing methods; (4) The last row shows the fitting results of the proposed MCC-VC-Laplacian method.} \label{f7a}
\end{figure}

To validate the fitting performance improvement by the proposed method, we select several images, each containing multiple ellipses. Fig. \ref{f6} shows the process of solving a multi-ellipse fitting problem with detection first and fitting. We first use Lu's detection method proposed recently in \cite{Lu2020} to detect the ellipses and then apply proposed MCC-VC-Laplacian methods to further fit the ellipses. We conducted corresponding comparative experiments, using the data points extracted after detection as input, using different methods, including the MCC-Laplacian\cite{hu2021}, HGMM\cite{HGMM}, SAREfit\cite{SARE2020}, for fitting. The results are shown in Fig. \ref{f7a}, where the threshold is set to 1.5. For comparison, the original detection results are also given in the second column. Although most ellipses are successfully detected by Lu's method, the fitting accuracy does not seem satisfactory, as seen from the second column. The proposed MCC-VC method has notable fitting performance improvement over Lu's method and it has the best fitting performance owing to its robustness. 

\begin{figure}[htb]
	\centering
	\includegraphics[height=2.5cm,width=5cm]{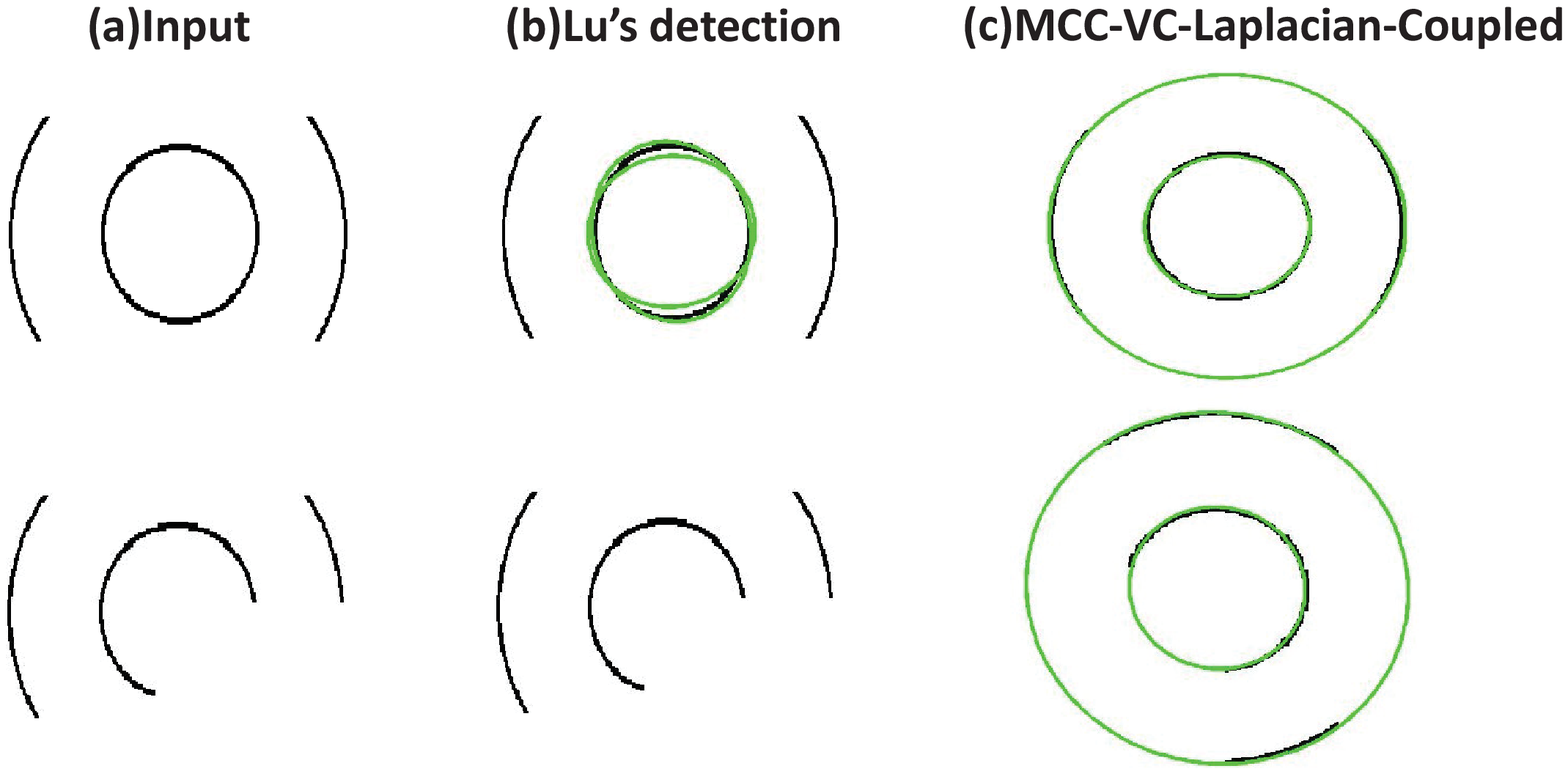}
	\caption{Illustration of the advantage of the proposed coupled ellipses fitting. (1) The first column of images are synthetic images; (2) The second column shows the detection results of Lu's method; (3) The last column shows the fitting results of the proposed coupled ellipses fitting method.} \label{f19}
\end{figure}

\begin{figure}[htb]
	\includegraphics[height=4.5cm,width=9cm]{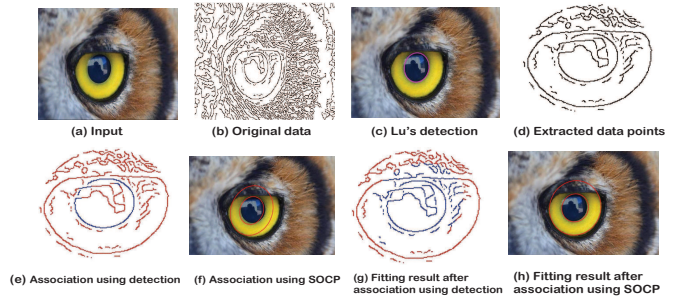}
	\caption{Illustration of the better performance of the proposed coupled ellipses fitting than Lu's method through the owl's iris image.} \label{f20}
\end{figure}

As a special case of multiple ellipses fitting, the coupled ellipses fitting can also be done through the procedure described above, where the association is accomplished based on ellipse detection. Nonetheless, it is straightforward to see that the fitting significantly depends on the detection performance since it follows after the detection. For ellipses lacking of arc segments, the detection method may fail, implying that the detection method cannot be used for grouping the data points.  In such a case, the proposed coupled ellipses fitting method offers better results when the image is known to contain coupled ellipses. Fig. \ref{f19} and Fig. \ref{f20} illustrate the advantage of the proposed coupled ellipses fitting through two synthetic images and a real image, respectively. It is seen from the second column of Fig. \ref{f19} that Lu's method cannot detect the ellipses owing to some losing portions. In comparison, the proposed coupled ellipses fitting method successfully fits the coupled ellipses. Fig. \ref{f20}(a) shows an owl's iris image containing coupled ellipses. For this image, Lu's detection method is only able to detect the inner ellipse; see the Fig. \ref{f20}(c). The data points associated to the inner ellipse can be determined using the detected inner ellipse, and the rest data points are associated to the outer ellipse. The association results based on detection and using the proposed SOCP method are given in the Fig. \ref{f20}(e)(f), respectively. The proposed MCC-VC coupled ellipses fitting method follows after the association, and the fitting results are shown in the Fig. \ref{f20}(g)(h), respectively. Clearly, the proposed method has better fitting performance. 

\section{Conclusion}\label{s6}
In this paper, we have presented a new ellipse fitting method based on the MCC-VC method, which offers strong robustness against outliers. By the iterative optimization of the kernel center, kernel bandwidth and ellipse parameter vector, the proposed MCC-VC method is more flexible and is applicable to more challenging scenarios. Furthermore, we have proposed a data association method for coupled ellipses fitting without knowing the data association between the ellipses and the data points, and extended the proposed MCC-VC method to coupled ellipses fitting.  Both simulated data and real images have confirmed the superior fitting performance of the proposed method over several recently proposed fitting methods. 

\begin{appendices}	
\section{The approximate convex function of $h(r)$}\label{app2}
By introducing a known positive constant $r_0$, we can rewrite $h(r)$ as 
\begin{small}
		\begin{align}\label{eabc}
			h(r)&=-\frac{r-r_0}{N}\sum^{N}_{i=1}e^{{-\hat{a}_{i}}r}+\frac{r}{4}-\frac{r_0}{N}\sum^{N}_{i=1}e^{{-\hat{a}_{i}}r}\nonumber\\
			&=h_1(r)+h_2(r),
		\end{align}	
\end{small}
where 
\begin{small}
		\begin{align}\label{eabcd}
			h_1(r)&=-\frac{r-r_0}{N}\sum^{N}_{i=1}e^{{-\hat{a}_{i}}r},\; h_2(r)&=\frac{r}{4}-\frac{r_0}{N}\sum^{N}_{i=1}e^{{-\hat{a}_{i}}r}. 
		\end{align}	
\end{small}
	
To approximate $h(r)$ by a convex function, we approximate the two non-convex functions $h_1(r)$ and $h_2(r)$ to convex functions through appropriate Taylor expansions, respectively. For the function $h_1(r)$, we perform the Taylor expansion to $e^{{-\hat{a}_{i}}r}$ up to the third order at $r_0$, giving
\begin{small}
		\begin{align}\label{e24}
			e^{{-\hat{a}_{i}}r}\approx& e^{{-\hat{a}_{i}}r_0}-\hat{a}_{i}e^{{-\hat{a}_{i}}r_0}(r-r_0)+\frac{\hat{a}_{i}^2}{2}e^{{-\hat{a}_{i}}r_0}(r-r_0)^2\nonumber\\
			&-\frac{\hat{a}_{i}^3}{6}e^{{-\hat{a}_{i}}r_0}(r-r_0)^3.
		\end{align}	
\end{small}
The value of $\hat{a}_{i}$ may be large in the presence of outliers, and keeping up to the third-order in the expansion is to guarantee a sufficiently accurate approximation. 
	
Substituting (\ref{e24}) into $h_1(r)$ yields an approximate function $f_1(r)$:
\begin{small}
		\begin{align}\label{e25}
			f_1(r)=&-\frac{r-r_0}{N}\sum^{N}_{i=1}[e^{{-\hat{a}_{i}}r_0}-\hat{a}_{i}e^{{-\hat{a}_{i}}r_0}(r-r_0)\nonumber\\
			&+\frac{\hat{a}_{i}^2}{2}e^{{-\hat{a}_{i}}r_0}(r-r_0)^2-\frac{\hat{a}_{i}^3}{6}e^{{-\hat{a}_{i}}r_0}(r-r_0)^3].
		\end{align}	
\end{small}
For notational simplicity, we further define
\begin{small}
		\begin{align}\label{e26}
			&b_1=\frac{1}{N}\sum_{i=1}^Ne^{{-\hat{a}_{i}}r_0},\;b_2=\frac{1}{N}\sum_{i=1}^N\hat{a}_{i}e^{{-\hat{a}_{i}}r_0},\nonumber\\
			&b_3=\frac{1}{N}\sum_{i=1}^N\hat{a}_{i}^2e^{{-\hat{a}_{i}}r_0},\;b_4=\frac{1}{N}\sum_{i=1}^N\hat{a}_{i}^3e^{{-\hat{a}_{i}}r_0},
		\end{align}	
\end{small}
which are known constants. Using these notations, $f_1(r)$ can be rewritten as
\begin{small}
		\begin{align}\label{e26a}
			f_1(r)=\frac{b_4}{6}(r-r_0)^4-\frac{b_3}{2}(r-r_0)^3+b_2(r-r_0)^2-b_1(r-r_0).
		\end{align}	
\end{small}		

Regarding the convexity of $f_1(r)$, we have the following proposition.
	
\textit{Proposition 1:} $f_1(r)$ is a strictly convex function in the domain $(0,+\infty)$.
	
\begin{proof} By letting $t = r-r_0$, we form a function $f_1(t)$. To prove the convexity of $f_1(r)$, we first prove that $f_1(t)$ is strictly convex.	$f_1(t)$ is a strictly convex function if and only if the second derivative is strictly greater than zero, i.e., $f_1''(t)>0$ \cite{convex}. It follows from the expression of $f_1(t)$ that
\begin{small}
  \begin{align}\label{e28}
   &f_1''(t)=2b_4t^2-3b_3t+2b_2,
  \end{align}	
\end{small}
which is a quadratic function. Since $b_4>0$,  $f_1''(t)$ is convex and has a minimum. The minimum value of $f_1''(t)$ is
		\begin{small}
			\begin{align}\label{e29}
				\min_{t>-r_0} \;f_1''(t)=\frac{b_{2}b_{4}-\frac{9}{16}b_{3}^2}{\frac{1}{2}b_4},
			\end{align}	
		\end{small}
which is reached at $t=\frac{3b_3}{4b_4}$.
		
Since $\frac{b_4}{2}>0$, we only need to show the numerator term  $b_{2}b_{4}-\frac{9}{16}b_{3}^2 > 0$. Using the expressions of $b_2,b_3,b_4$ in (\ref{e26}), 
\begin{small}
	\begin{align}\label{e30}
    &b_{2}b_{4}-\frac{9}{16}b_{3}^2=\nonumber\\
	&\frac{1}{N^2}\sum_{i=1}^N(\hat{a}_{i}e^{{-\hat{a}_{i}}r_0})\cdot\sum_{i=1}^N(\hat{a}_{i}^3e^{{-\hat{a}_{i}}r_0})-\frac{9}{16N^2}\left(\sum_{i=1}^N\hat{a}_{i}^2e^{{-\hat{a}_{i}}r_0}\right)^2.
    \end{align}	
\end{small}
According to the Cauchy-Schwarz inequality, we have
\begin{small}
			\begin{align}\label{e31}
				&\sum_{i=1}^N(\hat{a}_{i}^{\frac{1}{2}}e^{{{-\frac{1}{2}\hat{a}_{i}}r_0}})^2\cdot\sum_{i=1}^N(\hat{a}_{i}^{\frac{3}{2}}e^{{{-\frac{1}{2}\hat{a}_{i}}r_0}})^2\nonumber\\
				&\geq\left(\sum_{i=1}^N(\hat{a}_{i}^{\frac{1}{2}}e^{{{-\frac{1}{2}\hat{a}_{i}}r_0}})\cdot(\hat{a}_{i}^{\frac{3}{2}}e^{{{-\frac{1}{2}\hat{a}_{i}}r_0}})\right)^2\nonumber\\
				&=\left(\sum_{i=1}^N\hat{a}_{i}^2e^{{-\hat{a}_{i}}r_0}\right)^2> \frac{9}{16}\left(\sum_{i=1}^N\hat{a}_{i}^2e^{{-\hat{a}_{i}}r_0}\right)^2.
			\end{align}	
\end{small}
		
The last inequality holds when $\left(\sum_{i=1}^N\hat{a}_{i}^2e^{{-\hat{a}_{i}}r_0}\right)^2\neq 0$, which obviously is the case since  $\hat{a}_{i}=|\hat{\delta_{i}}-\hat{c}|$, and $\hat \delta_{i}$ $(i=1,$\ldots$, N)$ are random samples and cannot have the same value of $\hat{c}$. We conclude from (\ref{e31}) that  $f_1''(t)> 0$ holds, indicating that $f_1(t)$ is a strictly convex function.   Since $f_1(r)$ is the composition of the convex function $f_1(t)$ and the affine function $t = r-r_0$, it is also strictly convex. 
	\end{proof}

	Next, we will focus on the approximation of function $h_2(r)$. For $e^{{-\hat{a}_{i}}r}$ in $h_2(r)$, we only perform the first-order Taylor expansion at $r_0$, since it is difficult to prove the convexity of the higher order approximations. By doing so, we obtain the approximate function $f_2(r)$ of $h_2(r)$:
	\begin{small}
		\begin{align}\label{e30a}
			f_2(r)=&\frac{r}{4}-\frac{r_0}{N}\sum^{N}_{i=1}[e^{{-\hat{a}_{i}}r_0}-\hat{a}_{i}e^{{-\hat{a}_{i}}r_0}(r-r_0)]\nonumber\\
			=&\frac{r}{4}-r_0b_1+r_0b_2(r-r_0),
		\end{align}	
	\end{small}
	which is an affine function of $r$ and thus convex.
	
	Finally, the approximate convex function of $h(r)$ is
	\begin{small}
		\begin{align}\label{e30ba}
			f(r)=&f_1(r)+f_2(r)\nonumber\\
			=&\frac{b_4}{6}(r-r_0)^4-\frac{b_3}{2}(r-r_0)^3+b_2(r-r_0)^2\nonumber\\
			&+(b_2r_0-b_1+\frac{1}{4})(r-r_0)+(\frac{1}{4}r_0-b_1r_0).
		\end{align}	
	\end{small}						
	\section{Closed-Form Solution of $f'(r)=0$}\label{app3}
	
	In this appendix, we show the existence and uniqueness of the real root of $f'(r)=0$, and give the expression of the closed-form solution. 
	
	Similar to Appendix \ref{app2}, we first obtain the function $f(t)$ by letting $t = r-r_0$. $f'(t)$ can be expressed as
	\begin{small}
		\begin{align}\label{e30bac}
			f'(t)=\frac{2b_4}{3}t^3-\frac{3b_3}{2}t^2+2b_2t+(b_2r_0-b_1+\frac{1}{4}).
		\end{align}	
	\end{small}
	By defining the following notations,
	\begin{small}
		\begin{align}\label{e30bacd}
			&d_1=\frac{2b_4}{3},\;d_2=-\frac{3b_3}{2},\;d_3=2b_2,\;d_4=(b_2r_0-b_1+\frac{1}{4}),\nonumber\\
			&p=\frac{3d_1d_3-d_2^2}{3d_1^2},\;q=\frac{27d_1^2d_4-9d_1d_2d_3+2d_2^3}{27d_1^3},
		\end{align}	
	\end{small}
	and according to the Cardano formula, $f'(t)$ can be equivalently expressed as $
	f'(t)=t^3+pt+q$.  By the discriminant of the root $\Delta=(\frac{q}{2})^2+(\frac{p}{3})^3$, we can verify the existence and uniqueness of the solution. Similar to proving the non-negativity of (\ref{e30}), we can validate the non-negativity of $p$, which means $\Delta>0$, and the equation $f'(t)=0$ has only one real root. The root can be expressed as 
	\begin{small}
		\begin{align}\label{root}
			t=\frac{-d_2-(k_1^{\frac{1}{3}}+k_2^{\frac{1}{3}})}{3d_1},
		\end{align}	
	\end{small}
	where
	\begin{small}
		\begin{align}\label{root2}
			&k_{1}=e_1d_2+3d_1\left[\frac{-e_2+(e_2^2-4e_1e_3)^{\frac{1}{2}}}{2}\right],\nonumber\\
			&k_{2}=e_1d_2+3d_1\left[\frac{-e_2-(e_2^2-4e_1e_3)^{\frac{1}{2}}}{2}\right],
		\end{align}	\end{small}
with $e_1=d_2^2-3d_1d_3$, $e_2=d_2d_3-9d_1d_4$, and $e_3=d_3^2-3d_2d_4$.Finally, the solution of the equation $f'(r)=0$ can be obtained from (\ref{root}) as follows:
	\begin{small}
		\begin{align}\label{root+r0}
			r=\frac{-d_2-(k_1^{\frac{1}{3}}+k_2^{\frac{1}{3}})}{3d_1}+r0.
		\end{align}	
	\end{small}		
\end{appendices}



\end{document}